\documentclass[letterpaper, 10 pt, conference]{ieeeconf}
\IEEEoverridecommandlockouts
\usepackage[T1]{fontenc}
\usepackage{textcomp}
\usepackage{times}
\usepackage{microtype}

\usepackage{amsmath}
\usepackage{amssymb}
\usepackage[table]{xcolor}
\usepackage{colortbl}
\usepackage{graphicx}
\usepackage{capt-of}
\usepackage{graphics}
\usepackage{epsfig}
\usepackage{booktabs}
\usepackage{multirow}
\usepackage{adjustbox}
\usepackage{siunitx}
\usepackage{mathtools}
\usepackage{wrapfig}
\usepackage{comment}
\usepackage{xspace}
\usepackage{etoolbox}
\usepackage[ruled,linesnumbered,noend]{algorithm2e}
\usepackage{tikz}
\usetikzlibrary{decorations.pathreplacing,arrows.meta}
\usepackage{cite}
\usepackage{hyperref}
\usepackage{stfloats}
\usepackage[percent]{overpic}
\usepackage[most]{tcolorbox}
\usepackage{array}
\usepackage{float}
\usepackage{placeins}

\makeatletter
\renewcommand{\@IEEEfigurecaptionsepspace}{\vskip 2pt\relax}
\let\paper@makecaption\@makecaption
\long\def\@makecaption#1#2{%
  \ifx\@captype\@IEEEtablestring
    \begingroup\centering\footnotesize
    #1\par\vskip 1pt
    {\scshape #2\par}%
    \endgroup\vskip 2pt
  \else
    \paper@makecaption{#1}{#2}%
  \fi}
\makeatother

\usepackage{xcolor}
\usepackage{pict2e}

\let\labelindent\relax
\usepackage[inline]{enumitem}
\usepackage[normalem]{ulem}

\hypersetup{
    hypertexnames=false,
    colorlinks=true,
    citecolor=black,
    linkcolor=black,
    filecolor=black,
    urlcolor=blue,
    pdftitle={Test-Time Motion Steering for Perceptive Humanoid Fall Recovery},
    pdfauthor={Osher Azulay, Cheng-Lin Hsieh, Stella X. Yu},
    pdfsubject={},
    pdfkeywords={},
    pdfcreator={},
    pdfproducer={},
    pdfpagemode=FullScreen,
}

\newcommand{\ve}[1]{\mathbf{#1}}
\newcommand{\sname}{RISE\xspace}
\definecolor{filmcharcoal}{RGB}{100,100,90}
\definecolor{filmhole}{RGB}{238,235,228}
\newcommand{\filmrail}{%
  \begingroup\setlength{\fboxsep}{0pt}%
  \colorbox{filmcharcoal}{\hbox to\linewidth{%
    \rule{0pt}{0.85mm}%
    \leaders\hbox{\hspace{0.8mm}\textcolor{filmhole}{\rule[0.17mm]{0.5mm}{0.5mm}}\hspace{0.8mm}}\hfill}}%
  \endgroup
}

\title{
Test-Time Motion Steering for Perceptive Humanoid Fall Recovery
}

\author{
Osher Azulay$^{*}$, Cheng-Lin Hsieh$^{*}$, and Stella X. Yu
\thanks{$^{*}$Equal contribution. All authors are with the University of Michigan, Ann Arbor.}
}

\tikzset{ph/.style={draw=black!50, dashed, fill=black!3, rounded corners=2pt,
  align=center, text=black!60, inner sep=3pt, font=\scriptsize}}

\begin{document}

\raggedbottom
\setlength{\belowdisplayskip}{2pt}
\setlength{\belowdisplayshortskip}{3pt}
\setlength{\abovedisplayskip}{2pt}
\setlength{\abovedisplayshortskip}{3pt}
\setlength{\parskip}{0pt}

\markboth{Submitted for Review}
{
Test-Time Motion Steering for Perceptive Humanoid Fall Recovery
}

\newcommand{\arxivTeaser}{%
\noindent\begin{minipage}{\textwidth}
\centering
\includegraphics[width=\linewidth]{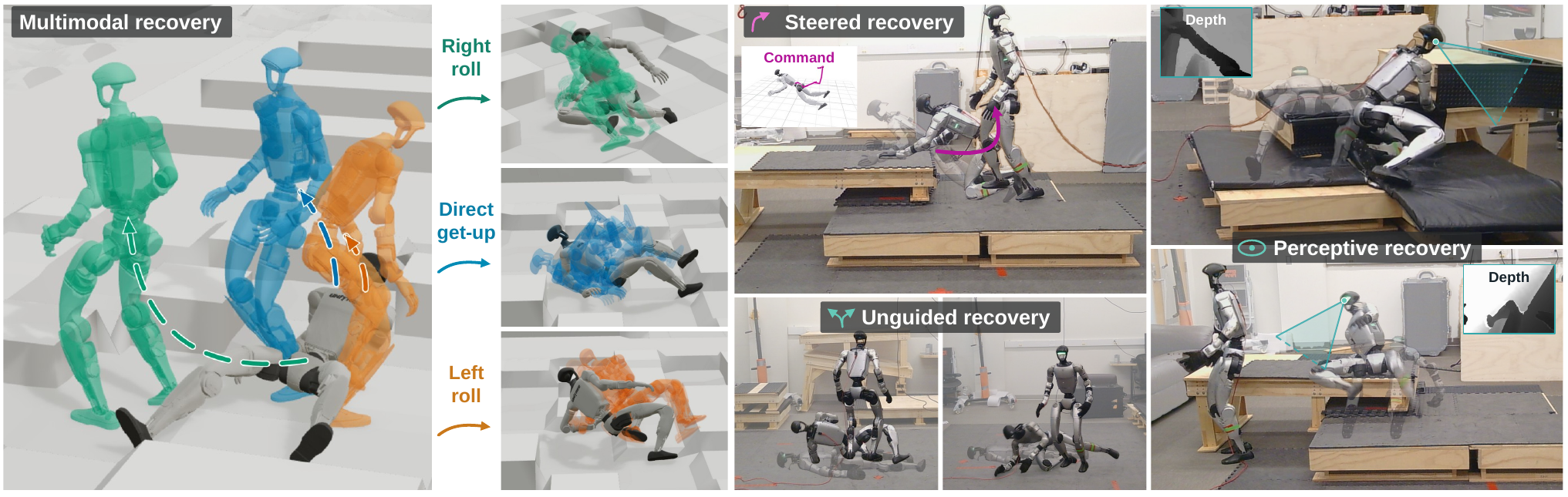}
\par\nointerlineskip
\captionof{figure}{\textbf{Perceptive and steerable humanoid fall recovery.}
From the same pose, the planner generates right-roll, direct get-up, and left-roll recoveries (left).
Real-world examples show steered and unguided recovery (middle) and visual adaptation to different support configurations (right).}
\label{fig:teaser}
\end{minipage}\par\vspace{4pt}%
}
\IEEEaftertitletext{\arxivTeaser}
\maketitle
\thispagestyle{empty}
\pagestyle{empty}

\begin{abstract}
Humanoid fall recovery requires selecting a motion that suits both the surrounding scene and the task. The same fallen pose can admit several ways to stand, with different routes, support transitions, and final headings. Yet reliable recovery alone does not provide control over which of these alternatives the robot executes. Steering this choice requires anticipating the physical consequences of a requested movement. We present \sname, an approach to perceptive humanoid recovery that lets task objectives steer how the robot returns to standing at inference time.
It builds on two insights: diverse recoveries can be composed from shared movements, and anticipating how these movements unfold allows recovery to be steered toward a task objective. We learn a recovery vocabulary that connects intended movements to motor actions through sensory feedback. A visual diffusion planner learns from executions of this vocabulary to predict motion intent as future body states. Task costs guide this prediction toward a desired heading, rise speed, or path. The planner then generates motion latents conditioned on the steered intent, allowing the requested movement to shape its execution. Simulation experiments demonstrate diverse recoveries, terrain adaptation, and test-time steering without retraining, complemented by real-world demonstrations on a Unitree G1. Project page: \href{https://osheraz.github.io/RISE/}{\texttt{RISE}}.
\end{abstract}

\section{Introduction}
\begin{figure}[t]
\centering
\includegraphics[width=\columnwidth,trim=0 5.5bp 0 4.5bp,clip]{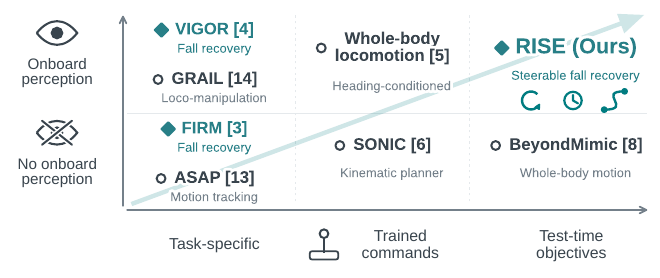}
\par\nointerlineskip
\caption{\textbf{Prior methods
offer onboard perception or test-time steering, not both.}
\textcolor[HTML]{287F87}{Diamonds} denote fall recovery;
\textcolor[HTML]{39444D}{circles} denote other whole-body motion.
\sname
combines both for fall recovery.}
\label{fig:literature-positioning}
\end{figure}

For a fallen humanoid, standing up is only part of recovery: the robot must also choose
\emph{how} to stand. Fig.~\ref{fig:teaser} shows that even the same fallen pose
admits several ways to stand, such as rolling to either side or rising
directly. The scene and next task
determine which fits: a wall can obstruct a recovery route, a step can
change the available support, and the task may require the robot to face a
particular direction. Steering lets task objectives select among feasible
recoveries, preparing the robot to continue its task.

Prior work has achieved reliable fall recovery and controllable whole-body
motion, but combining onboard perception with test-time steering remains under-explored, as Fig.~\ref{fig:literature-positioning} shows. Recovery
controllers now stand up from diverse postures and manage falls
\mbox{\cite{HuangT-RSS-25,HeX-RSS-25,xu2025unified}}, and VIGOR adapts
recovery to uneven ground using egocentric depth~\cite{azulay2026vigor}.
In these methods, the learned policy alone decides how the robot stands.
Other whole-body controllers,
including perceptive ones, are often directed through commands such as
joystick or heading inputs~\cite{zhang2026wholebody,luo2026sonic}. These
commands work well for trained behaviors, but each must be defined, with
labeled data, before training. Guided motion generation is more flexible: it
expresses task objectives as test-time costs that can be designed after
training and combined as needed~\cite{zhao2025dartcontrol,liao2026beyondmimic}.
We bring this flexibility to perceptive fall recovery.

Steering fall recovery is harder than steering from an upright pose. The
robot starts from varied fallen configurations and must shift support among
its hands, knees, feet, and torso to rise, so its motor actions must adapt to
its configuration and contacts. Uneven terrain compounds the difficulty:
steps, slopes, and nearby obstacles change the available supports and routes,
so a request feasible in one scene can fail in another. Yet this is the
practical setting, as humanoids rarely operate on flat ground with clear
space. Steering here needs several feasible ways to stand from each pose, but
each recorded recovery shows only one. It also needs to anticipate how a
requested movement will unfold under changing contact with the surrounding
terrain.

\begin{figure*}[t]
\centering
\includegraphics[width=\textwidth]{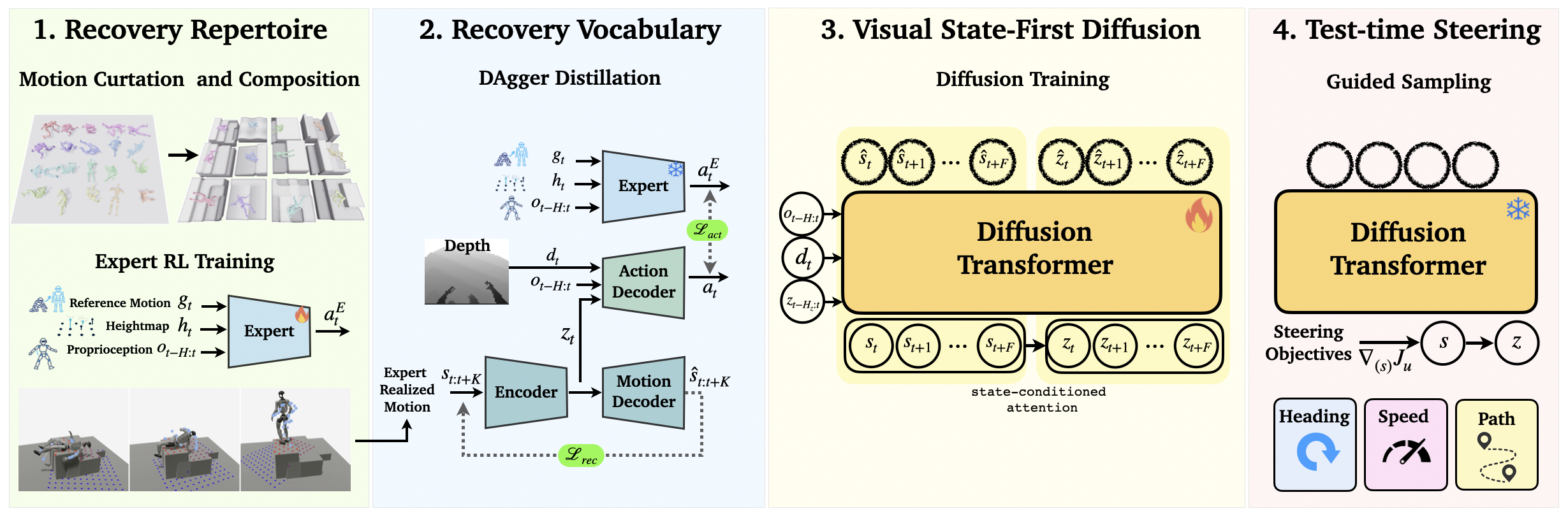}
\par\nointerlineskip
\caption{\textbf{From recorded recoveries to guided execution.}
(1) Composed motions are coarsely aligned to terrain for expert training.
(2) Expert supervision trains a recovery vocabulary and feedback decoder.
(3) Decoder executions pair motion latents with realized body states to train
the visual planner. (4) Task costs steer predicted motion intent, which then
conditions motion-latent generation for feedback execution.}
\label{fig:overview}
\end{figure*}

We introduce \sname (Perceptive \underline{R}ecovery via \underline{I}ntent \underline{S}t\underline{e}ering),
which addresses these needs with two insights (Fig.~\ref{fig:overview}).
\textit{First, diverse recoveries can be composed from shared movement
patterns}, providing several ways to stand from each pose. We compose
compatible recorded segments and adapt them to terrain, building a
\emph{multimodal} recovery repertoire. From the repertoire's physical
executions, we learn a \emph{recovery vocabulary}: motion latents that a
feedback decoder turns into motor actions using proprioception and egocentric
depth. \textit{Second, a planner trained on the robot's own executions can
anticipate how recovery will unfold}, so steering its predictions shapes the
executed recovery. Our visual diffusion planner
predicts this unfolding as \textit{motion intent}: a sequence of
future body states. Prior planners denoise future
states jointly with actions or motion
latents~\cite{huang2025diffusecloc,liao2026beyondmimic}. Our planner is
instead state-first: task costs steer the predicted states, then motion latents
are generated conditioned on them, making the requested recovery an explicit
condition for execution.

In
simulation, \sname recovers more reliably and with lower contact forces
than prior work while offering several ways to stand from the same pose.
A single trained model follows heading, speed, and path objectives across
terrains without retraining.
Planning with egocentric depth improves recovery even when the decoder already
uses depth. The state-first design tracks these objectives more accurately than
joint generation of states and motion latents.
Real-world demonstrations show perceptive and steerable recovery.

Our contributions are threefold:
\begin{enumerate}
[label=\arabic*),font=\bfseries,leftmargin=*,itemsep=4pt,topsep=0pt]
  \item We
  introduce \sname, to our knowledge the first framework for humanoid fall recovery
  that combines perception with test-time steering by task objectives.
  \item We propose a motion-matching pipeline that builds a multimodal
  recovery repertoire by composing recorded motions, coarsely adapting them
  to terrain, and collecting their physical executions.
  \item We introduce a visual state-first diffusion planner that steers
  predicted body states toward task objectives, then generates motion latents
  conditioned on those states to realize the requested recovery.
\end{enumerate}

\section{Related Work}
\label{sec:related}
\textbf{Humanoid Fall Recovery.}
Prior fall-recovery methods differ in how they design or learn recovery but
share one goal: returning the robot upright. Early work designed protective
falls and scripted get-up
sequences~\cite{fujiwara2002ukemi,stuckler2006getting}. Learned controllers
extended these capabilities to safe falling~\cite{kumar2017learning} and to
recovery from diverse postures on real
humanoids~\cite{HuangT-RSS-25,HeX-RSS-25}. FIRM unifies fall prevention,
impact mitigation, and recovery~\cite{xu2025unified}, and VIGOR uses
egocentric depth to infer a recovery goal suited to the nearby
terrain~\cite{azulay2026vigor}. None lets a task decide how the robot stands;
in \sname, task objectives choose the heading, speed, and path of
recovery at test time.

\textbf{Steering Whole-Body Motion.}
To our knowledge, no prior whole-body controller offers both onboard scene
perception and test-time steering~(Fig.~\ref{fig:literature-positioning}).
Many controllers, including perceptive ones, are directed through commands
such as joystick or heading inputs~\cite{luo2026sonic,zhang2026wholebody}.
These commands work well for trained behaviors, but each command type must be
defined, and its data labeled, before training. Test-time objectives are more
flexible: they can be designed after training and combined as needed. Such
objectives already steer simulated characters, whose state and scene are known
exactly~\cite{zhao2025dartcontrol,huang2025diffusecloc}. A real robot is
harder to steer: objectives act on predicted motion, but sensing, state
estimation, and mismatched dynamics~\cite{he2025asap} make the robot depart
from its plan. BeyondMimic steers a real humanoid, but on flat ground, and
notes that its guidance suits coarse-grained
objectives~\cite{liao2026beyondmimic}. Fall recovery on uneven terrain demands
finer control: contact-rich movement there is hard to predict, and far fewer
motions keep the robot supported. The robot must also stand up from arbitrary
initial configurations rather than a fixed starting state, sensing its joint
configuration and pelvis orientation but not where it lies on the terrain. We
steer recovery with egocentric depth, through predictions learned from the
robot's executed recoveries~(\mbox{Sec.~\ref{sec:eval-prediction}}).

\textbf{Multimodal Recovery Data.}
Steering requires several feasible recoveries from the same pose and position,
which recorded recovery data rarely provides. Such data is scarce next to the
motion data behind general whole-body control and
loco-manipulation~\cite{luo2026sonic,xie2026grail}. Large motion-capture
datasets include get-up demonstrations on flat
ground~\mbox{\cite{harvey2020robust,bones2026seed,mahmood2019amass}}, but because
recovery is not a focus category, they contain only a limited number.
Recovery controllers that imitate humans likewise rely on sparse
demonstrations~\cite{xu2025unified,azulay2026vigor}. Each recording also
shows only one way to stand from its start, since a performer can hardly fall
into the same pose twice. VIGOR aligns each demonstration to many terrains but
still pairs each starting pose with one recovery, repeating one route per
configuration in our tests~(\mbox{Sec.~\ref{sec:eval-safety}}). Motion
graphs~\cite{kovar2002motiongraphs}, physics-based
augmentation~\cite{xu2025parc}, and motion matching~\cite{wu2026perceptive}
broaden the skills and terrains that data covers. We use motion matching with
a different aim: composing recordings at compatible frames so that each
starting pose has several physically executable recoveries for steering
to choose among, rather than one.

\textbf{State--Action Diffusion Planning.}
Diffusion planners that output actions differ in when actions are generated
relative to the states they realize. Diffuse-CLoC and BeyondMimic denoise
states jointly with actions or motion
latents~\cite{huang2025diffusecloc,liao2026beyondmimic}. Under guidance, the
task cost changes the states at every denoising step while the actions are
updated alongside them. The actions are therefore never conditioned on the
final steered states, and whether they realize those states depends on the
shared denoising keeping the two consistent. We hypothesize that generating
actions from the completed steered plan helps the robot follow task
objectives while still completing recovery. Decision Diffuser follows this
ordering in simulation, inferring each action from consecutive states through
inverse dynamics~\cite{ajay2023decisiondiffuser}. Our planner instead
generates motion latents conditioned on the full steered state sequence, which a
feedback decoder executes on the real robot; \mbox{Sec.~\ref{sec:eval-joint}} compares
this ordering with joint generation.

\section{Method}
\begin{figure}[t]
\centering
\includegraphics[width=0.92\columnwidth,trim=0bp 32bp 0bp 28bp,clip]{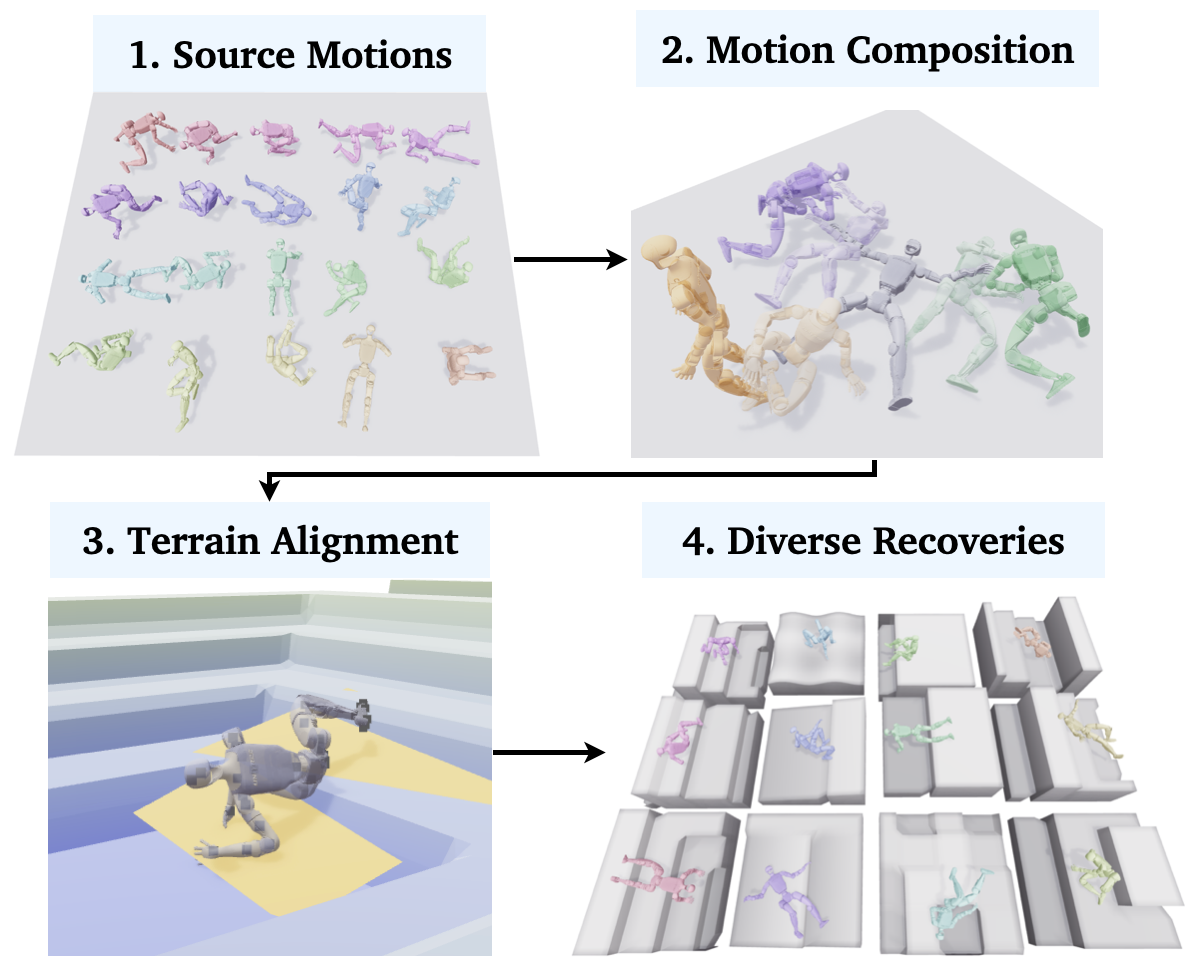}
\par\nointerlineskip
\caption{\textbf{Curating a multimodal recovery set.}
Recorded motions (1) are composed (2) and coarsely aligned to terrain (3),
yielding alternatives per starting pose (4) that a privileged expert learns
to execute in simulation.}
\label{fig:curation}
\end{figure}

\sname lets a fallen humanoid choose among different ways to get up and
steer its recovery toward a requested heading, rise speed, or path
(Fig.~\ref{fig:overview}). We first curate a \emph{multimodal recovery repertoire}.
We then learn a \emph{recovery vocabulary} and a
feedback decoder that turns the desired movement into motor actions. To \emph{plan executable recovery}, we train a planner on
the robot's resulting motions. Finally,
\emph{steering recovery with motion objectives} lets tasks shape predicted
motion intent and its execution.

\subsection{Curating a Multimodal Recovery Repertoire}
\label{sec:expert}

The curated recovery dataset supplies examples for learning the
recovery vocabulary (Fig.~\ref{fig:curation}).
We compose alternative recovery references, adapt them to terrain, and
verify physical feasibility with an expert policy.

\textbf{Motion composition.}
We create different ways to get up by combining compatible parts of recorded
recoveries. Our sources are
LAFAN1~\cite{harvey2020robust}, BONES-SEED~\cite{bones2026seed},
AMASS~\cite{mahmood2019amass}, and our recordings in G1 coordinates.
We use frame-level motion matching to compose variable-length recovery
segments. For an outgoing
frame $i$, we select a compatible frame from another recording,
$j^\star=\arg\min_{j\in\mathcal C(i)}\|\ve f_i-\ve f_j\|_2^2$,
where $\ve f$ contains normalized pose, velocity, body-attitude, and
limb-proximity features, and $\mathcal C(i)$ restricts candidates to compatible
frames near motion transitions. We align the recordings in position
and heading, blend the junction, and continue along the incoming recording.
Segment lengths follow the selected junctions rather than a fixed duration.
Retaining several compatible matches creates alternative recovery sequences.

\textbf{Terrain adaptation and expert realization.}
We align each reference with the ground to provide plausible support and
clearance for learning physical recovery. Building on terrain-based vertical
alignment~\cite{tessler2024maskedmimic,azulay2026vigor}, we add a local
support-plane rotation that tilts the fallen body
and fades during the rise. At each frame, we transform body points as
\begin{equation}
 \tilde{\ve p}=\ve R(\ve p-\ve c)+\ve c+\Delta z\,\ve e_z,
 \label{eq:terrainfit}
\end{equation}
where $\ve p$ and $\tilde{\ve p}$ are the original
and aligned points, $\ve c$ is the fitted plane's center, $\ve R$ provides
the tilt, and $\Delta z$ adds clearance along the vertical unit vector $\ve e_z$.
Geometric checks reject invalid support, penetration, and discontinuities.
Each task combines terrain, an initial pose, and a coarse reference.

\textbf{Expert training.} An expert policy $\pi_E$ learns to turn these
references into physical recoveries. Coarse references guide the recovery
maneuver while reinforcement learning
adapts its execution to terrain and contact.
We train the expert with PPO~\cite{ppo}
to output joint targets for PD control,
$\ve a_t^E=\pi_E(\ve o_{t-H:t},\ve g_t,\ve h_t)$,
where $\ve o_{t-H:t}$ is a history of $H$ proprioceptive observations,
$\ve g_t$ contains sparse reference targets, and $\ve h_t$ contains privileged
local terrain heights. Each observation $\ve o_t$ includes joint positions and
velocities, body angular velocity, gravity direction, and the previous motor action.
Expert execution
verifies feasibility and pairs realized recoveries with their coarse references.

\subsection{Learning a Recovery Vocabulary}
\label{sec:actionvae}

The recovery vocabulary connects short movements from the expert's recoveries
to a decoder that adjusts its motor actions to the robot's current
posture and nearby terrain.

\textbf{Motion representation.}
Motion intent comprises future body states encoding
joint configuration and pelvis motion:
\begin{equation}
 \ve s=[\ve q-\ve q_{\rm default},\dot{\ve q},\Delta\ve p,\ve r],
 \label{eq:motiontoken}
\end{equation}
where $\ve q$ and $\dot{\ve q}$ are joint angles
and velocities, $\ve q_{\rm default}$ is the default joint configuration,
$\Delta\ve p$ is pelvis displacement, and $\ve r$ stores the first two columns
of the pelvis rotation matrix. We anchor recorded poses to
the robot's current pelvis:
\begin{equation}
 \begin{aligned}
 \Delta\ve p_{t+i}&=(\ve R_t^h)^\top(\ve p^{\rm rec}_{t+i}-\ve p_t),\\
 \ve r_{t+i}&=\operatorname{Rot6D}((\ve R_t^h)^\top\ve R^{\rm rec}_{t+i}),
 \end{aligned}
 \label{eq:pelvisframe}
\end{equation}
where $t$ is the current time and $i$ the sample
offset. The superscript $\mathrm{rec}$ denotes recorded expert poses;
$\ve p_t$ and $\ve R_t^h$ are the current pelvis position and
gravity-aligned heading frame. $\operatorname{Rot6D}$ extracts the
first two rotation-matrix columns.

\textbf{Intent encoding and feedback.}
An encoder compresses the desired state window into
a motion latent. The decoder combines it with observations to produce
motor actions:
\begin{align}
 \ve z_t
 &=\mathcal E(\ve s_{t:t+K},\ve m),
 \label{eq:encoder}\\
 \hat{\ve a}_t
 &=\pi(\ve o_{t-H:t},\ve z_t,\ve d_t),
 \label{eq:actiondecoder}
\end{align}
where $K$ is the reference-window lookahead,
$\ve m$ marks available samples, $\ve z_t$ is the motion latent,
$\hat{\ve a}_t$ is the predicted motor action, and $\ve d_t$ is egocentric
depth.
Proprioception enables correction, and depth adapts actions to nearby
terrain.

\textbf{Imitation and reconstruction.}
Desired state windows come from recorded expert
executions, re-expressed relative to the robot's current pelvis position and
heading using~\eqref{eq:pelvisframe}.

We train the decoder on corrections to the robot's own movements using online
DAgger~\cite{dagger}. Corrective expert motor actions $\ve a_t^E$ are queried
at states visited by the decoder, using the recording's
original coarse reference at the corresponding reference time.
Deviations during decoder execution create varied intent--state pairings,
training the decoder to adapt recorded intent to states it encounters.

To encourage motion latents to retain the intended
movement, we reconstruct the desired state window from the latent alone:
$\hat{\ve s}_{t:t+K}=\mathcal R(\ve z_t)$, where $\mathcal R$ is a training-only
motion decoder.
The combined loss is
\begin{equation}
 \mathcal L=\operatorname{MSE}(\hat{\ve a}_t,\ve a_t^E)
 +\lambda\mathcal L_{\rm rec},
 \label{eq:actionloss}
\end{equation}
where $\mathcal L_{\rm rec}$ is the weighted mean
squared error between reconstructed and desired states over valid samples,
and $\lambda$ sets its weight. The loss weights normalized joint and pelvis
errors separately, emphasizing near-future samples.
\subsection{Learning to Plan Executable Recovery}
\label{sec:planner}

\textbf{Offline training data.}
The planner learns from the robot's realized motion under varied physical
and terrain conditions. We freeze the learned encoder and decoder and collect
rollouts using recorded motion intent across the curated tasks.
Pairing each motion latent used for control with
the resulting next state provides offline training sequences.

\textbf{Factorized, state-first diffusion.}
The planner predicts future body states before
generating the motion latents to realize them. Both predictions use the context
\begin{equation}
 \ve c_t=(\ve o_{t-H:t},\ve z_{t-H_z:t},\ve d_t),
 \label{eq:plannercondition}
\end{equation}
where $H_z$ is the number of preceding motion
latents. Future states share the current anchor in~\eqref{eq:pelvisframe}.
To steer motion intent while coupling it to execution, we factor
\begin{equation}
 p(\ve S,\ve Z\mid\ve c_t)
 =p(\ve S\mid\ve c_t)\,p(\ve Z\mid\ve S,\ve c_t),
 \label{eq:factorization}
\end{equation}
where $\ve S=\ve s_{t:t+F}$ contains the predicted
states and $\ve Z=\ve z_{t:t+F}$ their associated motion latents over $F$ future
steps. Steered motion intent $\ve S$ conditions latent generation.
This learned distribution forms our recovery prior.

A shared Transformer implements both conditional
distributions with diffusion~\cite{ho2020ddpm}.
State and latent tokens use separate input
projections and output heads. Directed attention allows state tokens to
attend to context and other states, while latent tokens additionally attend
to the complete state sequence. At inference, we first denoise future states
from Gaussian noise, then hold the resulting sequence fixed while denoising
the motion latents that realize it.

We train on paired latent and realized state sequences,
sampling state prediction and conditional latent
prediction with equal probability. For state prediction, we corrupt future
states and mask future latents; for latent prediction, we corrupt latents
and condition on the corresponding states. Both tasks
predict clean targets using weighted mean-squared error. Small, temporally
correlated perturbations of the conditioning states encourage tolerance to
prediction errors at deployment. Training requires no task-specific labels.

\subsection{Steering Recovery with Motion Objectives}
\label{sec:deployment-steering}

\textbf{Steering motion intent.}
A differentiable task cost on predicted motion intent enables
test-time steering without task-specific training pairs. During DDIM
sampling~\cite{song2021ddim}, we evaluate this cost on the denoised
prediction and backpropagate through the denoiser to compute guidance:
\begin{equation}
\Delta\ve x=-\eta\nabla_{\ve x}J\bigl(\ve S(\ve x)\bigr),
\label{eq:plannerguidance}
\end{equation}
where $\ve x$ is the noisy state sequence, $\ve S(\ve x)$ is its
denoised prediction, $J$ is the task cost, and $\eta$ is the effective guidance
step size.
This update shifts sampling toward lower task cost while the learned
prior supplies the recovery repertoire.

\textbf{From intent to execution.}
\looseness=-1
We hold the steered motion intent $\ve S^\star$ fixed while sampling
$\ve Z\sim p(\ve Z\mid\ve S^\star,\ve c_t)$, then decode them
into motor actions using current observations (Fig.~\ref{fig:overview}).
We execute a short prefix and replan with fresh observations.

We illustrate this steering approach with three objectives that steer recovery heading,
speed, and path.

\textbf{Heading steering.}
We steer recovery toward a desired heading by penalizing the
wrapped angular error:
\begin{equation}
 J_{\rm heading}=\operatorname{wrap}(\alpha_F-\alpha^\star)^2,
 \label{eq:yawcost}
\end{equation}
where $\alpha_F$ is the predicted heading at the final step $F$,
$\alpha^\star$ is the target heading, and $\operatorname{wrap}$ maps angles
to $[-\pi,\pi)$.

\textbf{Speed steering.}
We steer how quickly the robot rises by guiding its height profile
over time:
\begin{equation}
 J_{\rm speed}=\frac1F\sum_{j=1}^F
 (h_j-h_j^\star)^2,
 \label{eq:heightcost}
\end{equation}
where $h_j$ and $h_j^\star$ are the predicted and desired pelvis
heights at future step $j$, both relative to the initial pelvis height.
Speed targets are expressed as rise times.

\begingroup\clubpenalty=10000
\textbf{Path steering.}
We guide the pelvis path and posture by matching target poses:
\begin{equation}
 J_{\rm path}=\frac1F\sum_{j=1}^F\!\left[
 \|\ve p_j-\ve p_j^\star\|_2^2
 +\lambda_R\|\ve R_j-\ve R_j^\star\|_F^2\right],
 \label{eq:posecost}
\end{equation}
where $\ve p_j$ and $\ve R_j$ are the predicted pelvis position and
rotation matrix at step $j$, and stars mark targets. $\lambda_R$ weights
orientation, and $\|\cdot\|_F$ is the Frobenius norm.
\par\endgroup

\section{Experiments}
\label{sec:experiments}
Humanoid recovery should remain reliable and safe across
scenes while offering diverse ways to stand. We evaluate
recovery quality and diversity (Sec.~\ref{sec:eval-safety}),
motion prediction accuracy (Sec.~\ref{sec:eval-prediction}), and
task-directed recovery without retraining (Sec.~\ref{sec:eval-steering}),
using the metrics in Table~\ref{tab:eval-metrics}. Further comparisons examine
visual feedback, prediction order, and replanning
(\mbox{Secs.~\ref{sec:eval-depth}--\ref{sec:eval-joint}}), and real-world
demonstrations illustrate recovery and steering
(Sec.~\ref{sec:realworld}).

\subsection{Experimental Setup}
\label{sec:eval-setup}\label{app:eval-contract}

\textbf{Simulation setup.}
We evaluate the 29-DoF Unitree G1 in Isaac Sim using our visual diffusion
planner and frozen depth-conditioned decoder (Sec.~\ref{sec:actionvae}). Unless
noted otherwise, we sample a one-second motion intent and its motion
latents at 50\,Hz using eight DDIM steps per pass, execute the first 0.2\,s,
and replan with fresh noise. Attempts last 15\,s. 

\begin{figure}[t]
\centering
\includegraphics[width=\columnwidth,trim=0 55 0 12,clip]{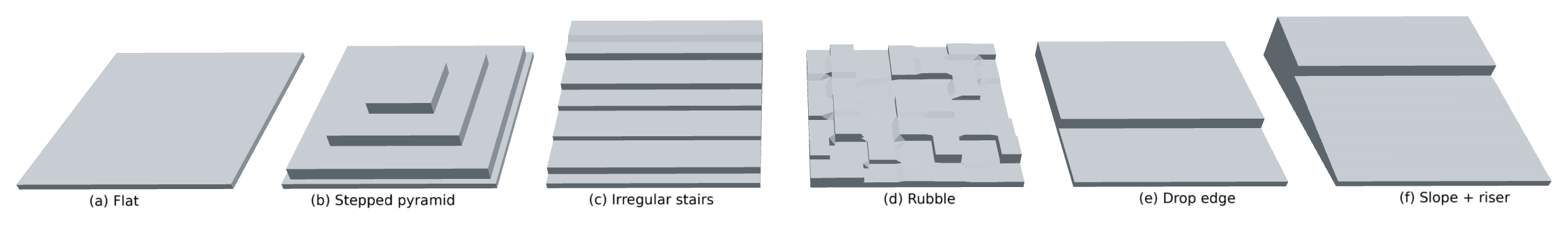}
\par
{\footnotesize\sffamily%
\makebox[\dimexpr\columnwidth/6\relax]{(a)}%
\makebox[\dimexpr\columnwidth/6\relax]{(b)}%
\makebox[\dimexpr\columnwidth/6\relax]{(c)}%
\makebox[\dimexpr\columnwidth/6\relax]{(d)}%
\makebox[\dimexpr\columnwidth/6\relax]{(e)}%
\makebox[\dimexpr\columnwidth/6\relax]{(f)}%
}
\par\nointerlineskip
\caption{\textbf{Evaluation terrains.} (a) Flat, (b) stepped pyramid,
(c) stairs, (d) rubble, (e) drop edge, and (f) slope with a riser.}
\label{fig:eval-terrain}
\end{figure}

\textbf{Environments and initial configurations.}
As in VIGOR~\cite{azulay2026vigor},
expert training pairs recovery references with sampled terrains
(Sec.~\ref{sec:expert}). We use this construction to sample
initial pose--terrain configurations held out from training.
The terrains include flat ground, stepped pyramids, stairs, rubble,
drop edges, and slopes with risers (Fig.~\ref{fig:eval-terrain}).

\textbf{Baselines.}
We compare fall-recovery performance against
VIGOR~\cite{azulay2026vigor} and FIRM~\cite{xu2025unified}, using checkpoints
provided by their authors. VIGOR uses visual feedback and acts
deterministically. FIRM samples
from its blind diffusion policy without added exploration noise. All methods use
the same initial configurations, recovery criterion, and time limit.
\emph{Joint diffusion} (Joint) denoises states and motion latents together.

\begin{table}[t]
\centering\footnotesize\setlength{\tabcolsep}{3pt}
\renewcommand{\arraystretch}{1.05}
\caption{Evaluation metrics}
\label{tab:eval-metrics}
\begin{tabular*}{\columnwidth}{@{\extracolsep{\fill}}ll@{}}
\toprule
\textbf{Metric} & \textbf{Definition} \\
\midrule
SR (\%) & Stable recovery through the end of the trial. \\
Pelvis displ. (m) & Mean final horizontal displacement. \\
Span (cm, deg) & 10th--90th percentile spread per configuration. \\
Peak force (kN) & Maximum net link contact force. \\
Peak torque (N\,m) & Largest absolute motor torque. \\
Saturation (\%) & Joint--time fraction at $\geq99\%$ torque limit. \\
Joint-limit count & Peak count of joints beyond limits by ${>}1$\,mrad. \\
Limit excess (mrad) & Largest excursion beyond a joint limit. \\
\bottomrule
\end{tabular*}
\end{table}

\subsection{Recovery Quality and Diversity}
\label{sec:eval-safety}\label{app:impl-a3}\label{sec:eval-multimodality}\label{app:impl-a1}

We test whether the learned prior offers several reliable ways to stand.
We compare all methods across 46 initial
configurations, with 20 trials each
(920 trials per method).

\begin{table}[t]
\centering\footnotesize\setlength{\tabcolsep}{2pt}
\renewcommand{\arraystretch}{1.05}
\caption{Unguided recovery comparison}
\label{tab:eval-e1}\label{tab:eval-e2}
\begin{tabular*}{\columnwidth}{@{\extracolsep{\fill}}llrrr@{}}
\toprule
\textbf{Category} & \textbf{Metric} & \textbf{FIRM}~\cite{xu2025unified} & \textbf{VIGOR}~\cite{azulay2026vigor} & \textbf{\sname} \\
\midrule
\textit{Performance} & SR (\%) $\uparrow$ & 8.2 & 80.8 & \textbf{93.0} \\
\midrule
\multirow{5}{*}{\shortstack[l]{\textit{Safety and}\\\textit{effort} $\downarrow$}} & Peak force (kN) & 6.5 & 2.5 & \textbf{1.4} \\
& Peak torque (N\,m) & 127.3 & 101.3 & \textbf{83.3} \\
& Saturation (\%) & 8.5 & 0.34 & \textbf{0.30} \\
& Joint-limit count & 0.8 & 0.6 & \textbf{0.2} \\
& Limit excess (mrad) & 6.7 & 1.7 & \textbf{1.4} \\
\midrule
\multirow{3}{*}{\shortstack[l]{\textit{Motion}\\\textit{characteristics}}} & Forward span (cm) & 37.8 & 0.0 & 32.9 \\
& Lateral span (cm) & 29.7 & 0.0 & 44.6 \\
& Heading span (deg) & 35.9 & 0.0 & 107.8 \\
\bottomrule
\end{tabular*}
\end{table}

\begin{table}[!t]
\centering\footnotesize
\setlength{\tabcolsep}{6pt}\renewcommand{\arraystretch}{1.05}
\caption{Prediction--execution errors}
\label{tab:eval-main-mechanism}
\begin{tabular*}{\columnwidth}{@{\extracolsep{\fill}}lccccc@{}}
\toprule
& \multicolumn{3}{c}{Pelvis} & \multicolumn{2}{c}{Limbs (L/R)} \\
\cmidrule(lr){2-4}\cmidrule(l){5-6}
Terrain & \shortstack{Heading\\(deg)} & \shortstack{Height\\(cm)} & \shortstack{Position\\(cm)} & \shortstack{Hands\\(cm)} & \shortstack{Soles\\(cm)} \\
\midrule
Flat & 3.6 & 1.5 & 3.1 & 6.1/5.8 & 6.4/6.5 \\
Uneven & 6.0 & 2.1 & 4.6 & 9.0/9.1 & 9.5/9.8 \\
\bottomrule
\end{tabular*}
\end{table}

\begin{figure}[!t]
\centering
\includegraphics[width=\columnwidth]{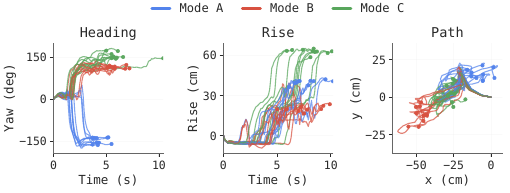}
\par\nointerlineskip
\caption{\textbf{Recovery diversity.} Each panel shows three recovery
patterns (modes) from one initial configuration, with ten successful
trials per mode.}
\label{fig:eval-multimodality}
\end{figure}

\textbf{Recovery quality.}
In Table~\ref{tab:eval-e1}, \sname improves success over both baselines
with lower contact forces, torques, and joint-limit violations, supporting
safer recovery.

\textbf{Multimodal recovery.}
Repeated recoveries from a fixed initial configuration produce
distinct trajectories, turning directions, and rise profiles
(Fig.~\ref{fig:eval-multimodality}). VIGOR repeats one route per
configuration, so its motion spans in Table~\ref{tab:eval-e2} are zero.
FIRM produces varied motions but rarely completes stable
recovery in these tests.
Our learned prior captures diverse and reliable
recovery strategies.

\subsection{Motion Prediction Accuracy}
\label{sec:eval-prediction}\label{app:impl-a4}

Task costs act on predicted motion intent. For these costs to
shape recovery, the prediction must reflect how the robot will move during
execution. We repeat 32 initial configurations twice on flat ground and
twice on uneven terrain (128 trials total). For this test, we replan every 0.4\,s and compare
predicted and executed pelvis, hand, and sole motion at matching times
within each interval.

Table~\ref{tab:eval-main-mechanism} shows close agreement in pelvis
heading, height, and position, with larger deviations at the hands and
soles, especially on uneven terrain. This supports steering body motion
through the predicted trajectory while allowing the decoder to
adjust limb motion to local support.

\subsection{Test-Time Steering of Recovery}
\label{sec:eval-steering}\label{app:impl-a2}

\begin{figure}[t]
\centering
\includegraphics[width=\columnwidth]{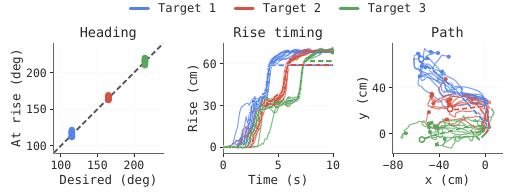}
\par\nointerlineskip
\caption{\textbf{Task-directed recovery.} Ten successful trials per target
from a fixed flat-ground configuration. Rise timing shows speed steering.
Dashed curves and open circles mark targets; the diagonal marks zero heading
error.}
\label{fig:eval-steering}
\end{figure}

Different tasks favor different ways of returning to standing.
We evaluate steering within the recovery modes captured by the learned
prior. Task costs guide heading, speed, or path without retraining
(Sec.~\ref{sec:deployment-steering}). We test on flat ground, stairs, and
a slope with a riser. For each objective and terrain, the two targets in
Table~\ref{tab:eval-flat-steering} share an initial configuration,
with 100 trials per target
(1,800 total). Paths A and B are two reference pelvis paths per terrain.
Fig.~\ref{fig:eval-steering} shows three targets per objective in a
separate qualitative comparison.

\begin{figure}[!t]
\centering
\begin{minipage}{\columnwidth}
\centering
\setlength{\fboxsep}{0.5pt}
\begin{tikzpicture}
\node[inner sep=0] (image) {\includegraphics[width=\columnwidth]{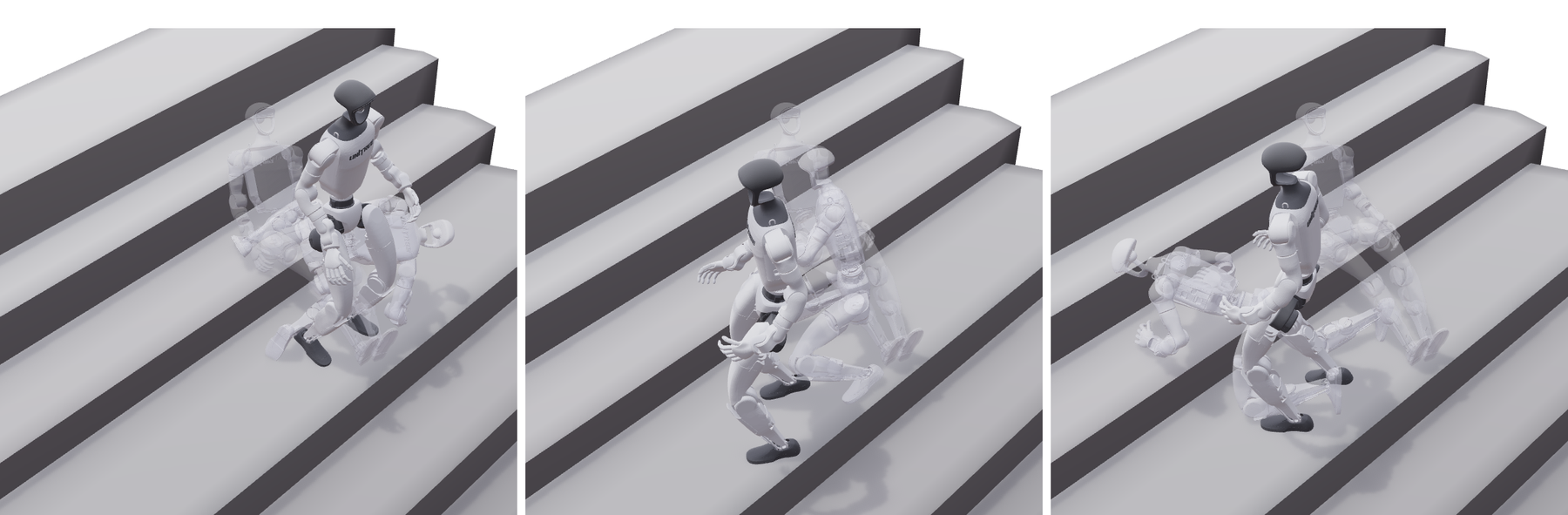}};
\node[anchor=north west,inner sep=0] at (image.north west)
{\colorbox[HTML]{DCE8F6}{\makebox[\dimexpr\columnwidth-2\fboxsep\relax]{\footnotesize\bfseries\textcolor[HTML]{24466B}{Unguided}}}};
\end{tikzpicture}
\par\vspace{0.5pt}\nointerlineskip
\begin{tikzpicture}
\node[inner sep=0] (image) {\includegraphics[width=\columnwidth]{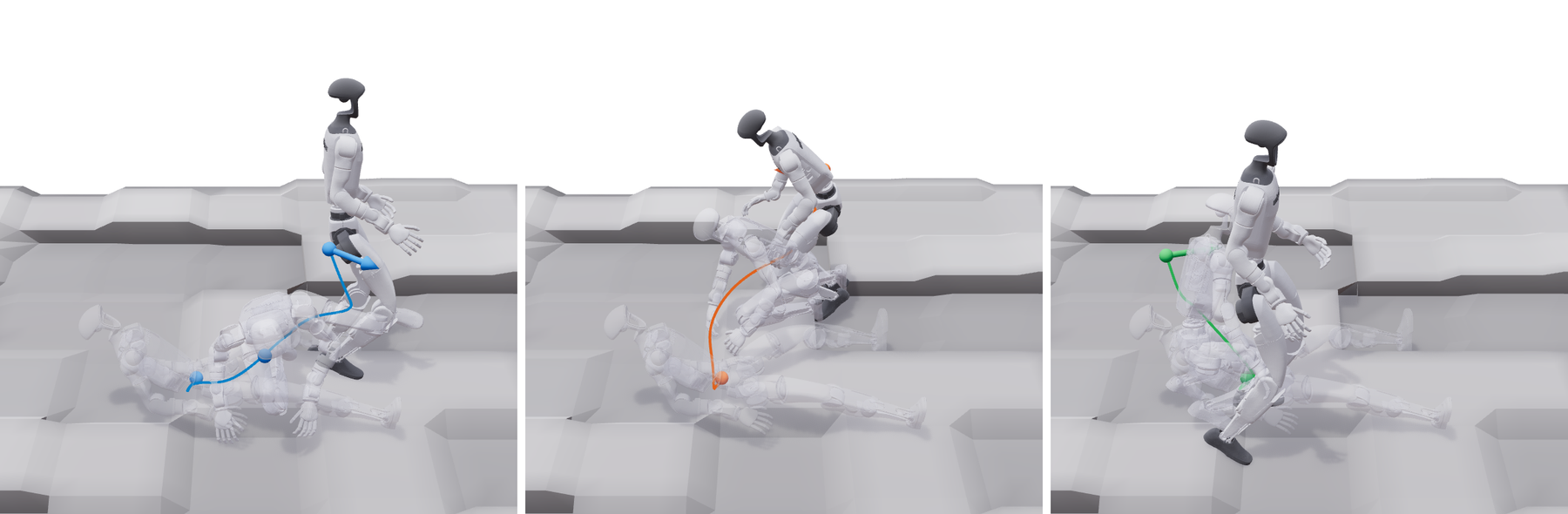}};
\node[anchor=north west,inner sep=0] at (image.north west)
{\colorbox[HTML]{DCEEDC}{\makebox[\dimexpr\columnwidth-2\fboxsep\relax]{\footnotesize\bfseries\textcolor[HTML]{275735}{Path steering}}}};
\end{tikzpicture}
\par\vspace{0.5pt}\nointerlineskip
\includegraphics[width=\columnwidth,trim=0bp 0bp 0bp 70bp,clip]{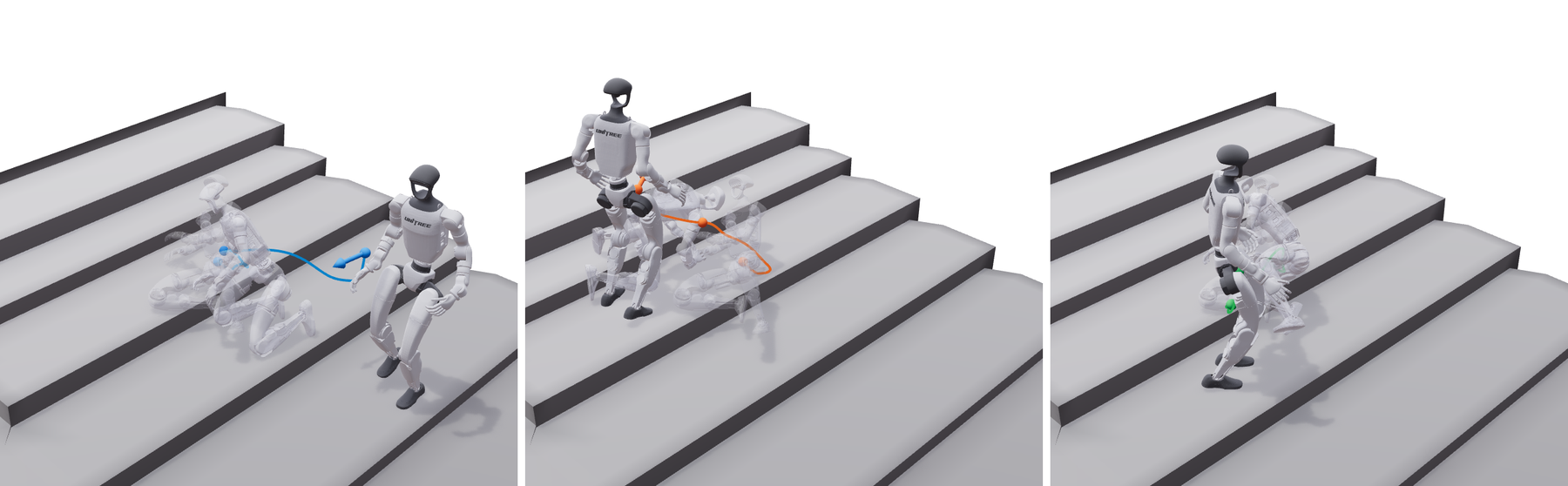}
\par\vspace{0.5pt}\nointerlineskip
\begin{minipage}[t]{0.415\linewidth}
\vspace{0pt}
\colorbox[HTML]{E9E0F4}{\makebox[\dimexpr\linewidth-2\fboxsep\relax]{\footnotesize\bfseries\textcolor[HTML]{5C3E7A}{Heading steering}}}\par\nointerlineskip
\includegraphics[width=\linewidth]{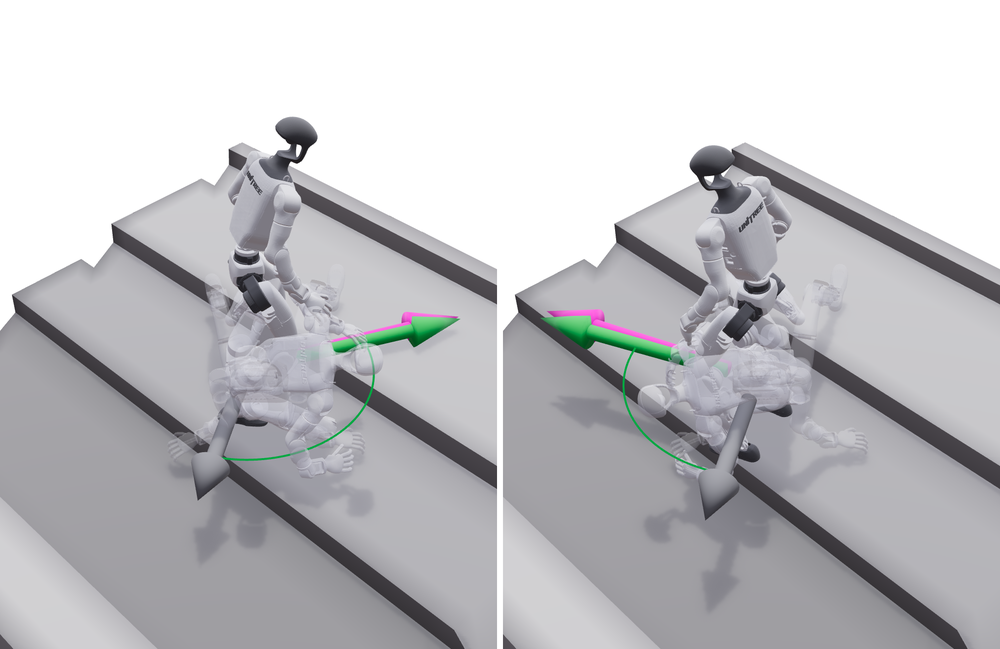}
\end{minipage}\hfill
\begin{minipage}[t]{0.570\linewidth}
\vspace{0pt}
\colorbox[HTML]{FBE3CD}{\makebox[\dimexpr\linewidth-2\fboxsep\relax]{\footnotesize\bfseries\textcolor[HTML]{854511}{Speed steering}}}\par\nointerlineskip
\begin{tikzpicture}[
timelabel/.style={font=\tiny,fill=white,text=black!85,inner xsep=0.7pt,
  inner ysep=0.45pt,rounded corners=0.4pt},
timepointer/.style={-{Latex[length=1.4pt,width=1.0pt]},
  draw=black!75,line width=0.25pt}]
\node[anchor=south west,inner sep=0] (timing) at (0,0)
  {\includegraphics[width=\linewidth]{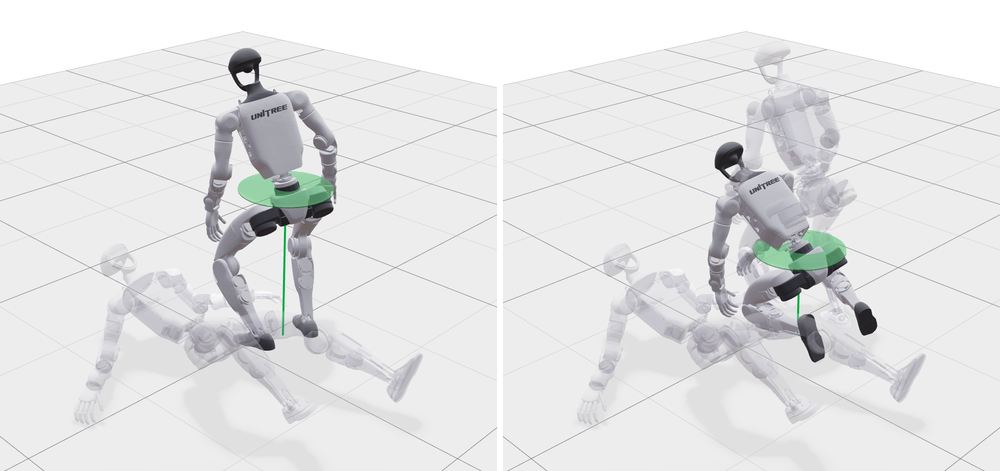}};
\begin{scope}[x={(timing.south east)},y={(timing.north west)}]
\node[timelabel] (early) at (0.13,0.77) {6.5\,s};
\node[timelabel] (late) at (0.61,0.59) {6.5\,s};
\node[timelabel] (standing) at (0.89,0.84) {8.0\,s};
\draw[timepointer] (early.east)--(0.222,0.82);
\draw[timepointer] (late.east)--(0.711,0.66);
\draw[timepointer] (standing.west)--(0.795,0.87);
\end{scope}
\end{tikzpicture}
\end{minipage}
\end{minipage}
\par\nointerlineskip
\caption{\textbf{Test-time steering of recovery path, heading, and speed.}}
\label{fig:sim-steering-examples}
\end{figure}

\begin{table}[!t]
\centering\footnotesize\setlength{\tabcolsep}{1.25pt}
\renewcommand{\arraystretch}{1.12}
\caption{Test-time steering results}
\label{tab:eval-flat-steering}
\begin{tabular*}{\columnwidth}{@{\extracolsep{\fill}}llccc@{\hspace{7pt}}ccc@{}}
\toprule
& & \multicolumn{3}{c}{Target 1} & \multicolumn{3}{c}{Target 2} \\
\cmidrule(lr){3-5}\cmidrule(l){6-8}
\textbf{Objective} & \textbf{Terrain} & Value & Error & SR (\%) & Value & Error & SR (\%) \\
\midrule
\multirow{3}{*}{Heading ($^\circ$)}
& Flat & 115.0 & 3.5 & 99.0 & 215.0 & 3.3 & 99.0 \\
& Stairs & 345.0 & 4.7 & 96.0 & 285.0 & 6.6 & 98.0 \\
& Slope + riser & 275.0 & 4.4 & 100.0 & 0.0 & 5.3 & 98.0 \\
\midrule
\multirow{3}{*}{\shortstack[l]{Speed\\(rise time, s)}}
& Flat & 6.0 & 0.3 & 97.0 & 7.5 & 0.1 & 99.0 \\
& Stairs & 3.5 & 0.3 & 90.0 & 4.2 & 0.03 & 100.0 \\
& Slope + riser & 4.8 & 0.7 & 99.0 & 5.7 & 0.6 & 100.0 \\
\midrule
\multirow{3}{*}{Path (cm)}
& Flat & Path A & 12.5 & 97.0 & Path B & 14.1 & 93.0 \\
& Stairs & Path A & 12.4 & 94.0 & Path B & 12.0 & 80.0 \\
& Slope + riser & Path A & 12.2 & 89.0 & Path B & 14.5 & 90.0 \\
\bottomrule
\end{tabular*}
\end{table}

\begin{table}[!t]
\centering\footnotesize\setlength{\tabcolsep}{4pt}
\caption{Visual feedback and recovery}
\label{tab:depth-perception}
\begin{tabular*}{\columnwidth}{@{\extracolsep{\fill}}lccrrrr@{}}
\toprule
\multirow{3}{*}{Variant} & \multicolumn{2}{c}{Depth} & \multirow{3}{*}{\shortstack{SR\\(\%)}} & Pelvis & Peak & Limit \\
\cmidrule(lr){2-3}
& \multirow{2}{*}{Planner} & \multirow{2}{*}{Decoder} & & displ. & torque & excess \\
& & & & (m) & (N\,m) & (mrad) \\
\midrule
Full model & $\checkmark$ & $\checkmark$ & \textbf{93.3} & \textbf{0.8} & \textbf{89.3} & \textbf{2.4} \\
Blind planner & $\times$ & $\checkmark$ & 85.0 & 0.9 & 93.6 & 3.4 \\
Both blind & $\times$ & $\times$ & 79.2 & 1.0 & 96.0 & 3.7 \\
\bottomrule
\end{tabular*}
\end{table}

\begin{table}[!t]
\centering\footnotesize\setlength{\tabcolsep}{3.5pt}
\caption{Prediction order and replanning interval}
\label{tab:steering-replanning}
\label{tab:eval-joint}
\begin{tabular*}{\columnwidth}{@{\extracolsep{\fill}}clrrrrrr@{}}
\toprule
& & \multicolumn{2}{c}{Heading} & \multicolumn{2}{c}{Speed} & \multicolumn{2}{c}{Path} \\
\cmidrule(lr){3-4}\cmidrule(lr){5-6}\cmidrule(l){7-8}
Replan & Generation & {SR} & {Error} & {SR} & {Error} & {SR} & {Error} \\
(s) & & {(\%)} & {(deg)} & {(\%)} & {(cm)} & {(\%)} & {(cm)} \\
\midrule
\multirow{2}{*}{0.2} & State-first & \textbf{90.0} & \textbf{5.3} & \textbf{94.2} & \textbf{1.8} & \textbf{89.2} & \textbf{16.3} \\
 & Joint & 89.2 & 7.1 & 93.3 & 3.0 & 78.3 & 23.1 \\
\addlinespace
\multirow{2}{*}{0.5} & State-first & \textbf{96.7} & \textbf{7.3} & 94.2 & \textbf{3.0} & \textbf{72.5} & \textbf{23.3} \\
 & Joint & 95.8 & 8.1 & \textbf{95.0} & 4.8 & 64.2 & 27.4 \\
\addlinespace
\multirow{2}{*}{1.0} & State-first & \textbf{99.2} & \textbf{13.5} & \textbf{99.2} & 6.1 & \textbf{85.8} & \textbf{26.1} \\
 & Joint & 76.7 & 13.8 & 94.2 & \textbf{6.0} & 83.3 & 29.2 \\
\bottomrule
\end{tabular*}
\par\vspace{1pt}
{\raggedright\scriptsize State-first denotes \sname (ours).\par}
\end{table}

\begin{figure*}[!t]
\centering
\begin{minipage}[t]{1\textwidth}
{\footnotesize\textbf{(A) Steered recovery across raised platforms}\hfill{\scriptsize Time $\longrightarrow$}}\par\vspace{1pt}\nointerlineskip
\begin{minipage}[c]{0.125\linewidth}
\centering
{\scriptsize\color{magenta!70!black} Guidance}\par
\includegraphics[width=\linewidth]{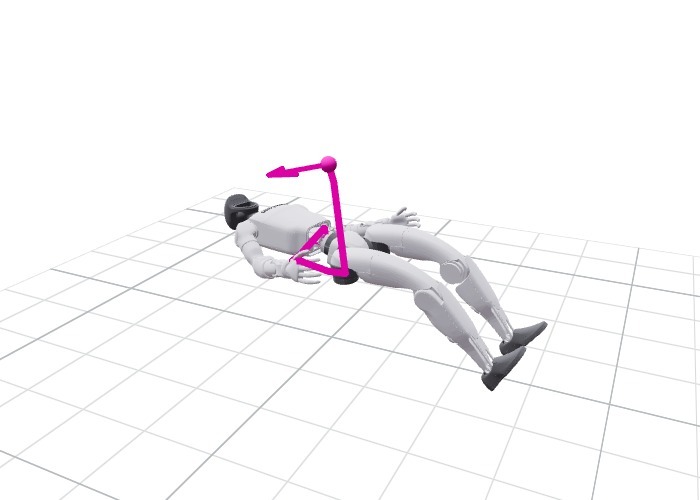}
\end{minipage}\hfill
\begin{minipage}[c]{0.865\linewidth}
\filmrail\par\nointerlineskip
\begin{minipage}[t]{0.247000\linewidth}
\centering
\begin{tikzpicture}
\node[inner sep=0,outer sep=0] (scene) {\includegraphics[width=\linewidth,trim=0bp 0bp 0bp 60bp,clip]{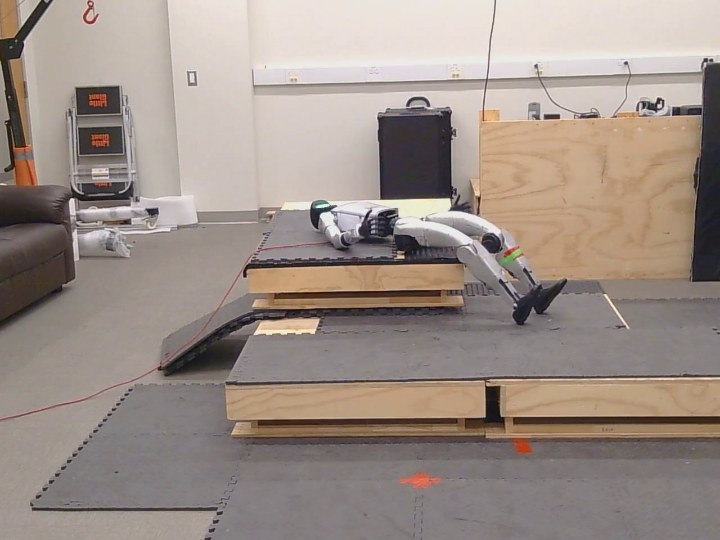}};
\node[anchor=north east,inner sep=0,outer sep=0,draw=white,line width=0.25pt] at (scene.north east) {\includegraphics[width=0.18623\linewidth,trim=465bp 341bp 0bp 0bp,clip]{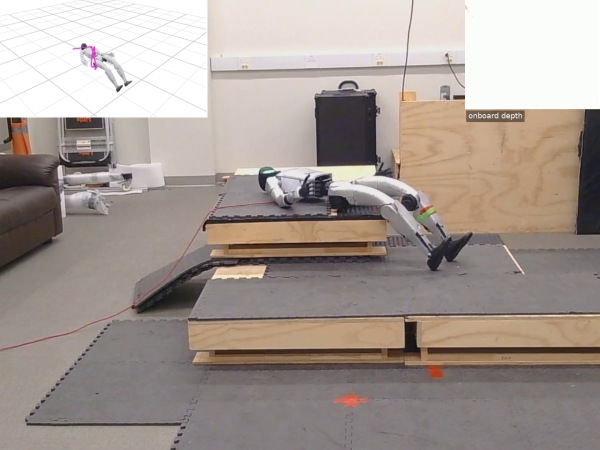}};
\end{tikzpicture}
\end{minipage}\hfill
\begin{minipage}[t]{0.247000\linewidth}
\centering
\begin{tikzpicture}
\node[inner sep=0,outer sep=0] (scene) {\includegraphics[width=\linewidth,trim=0bp 0bp 0bp 60bp,clip]{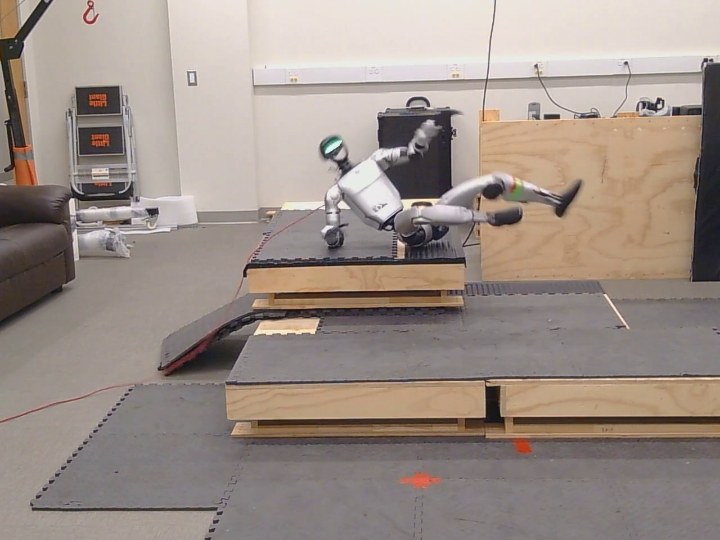}};
\node[anchor=north east,inner sep=0,outer sep=0,draw=white,line width=0.25pt] at (scene.north east) {\includegraphics[width=0.18623\linewidth,trim=465bp 341bp 0bp 0bp,clip]{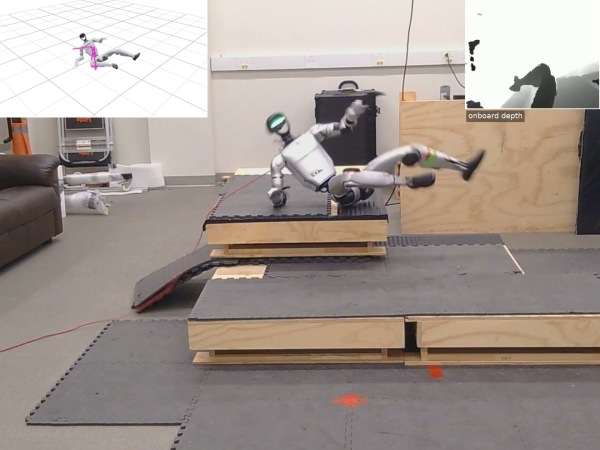}};
\end{tikzpicture}
\end{minipage}\hfill
\begin{minipage}[t]{0.247000\linewidth}
\centering
\begin{tikzpicture}
\node[inner sep=0,outer sep=0] (scene) {\includegraphics[width=\linewidth,trim=0bp 0bp 0bp 60bp,clip]{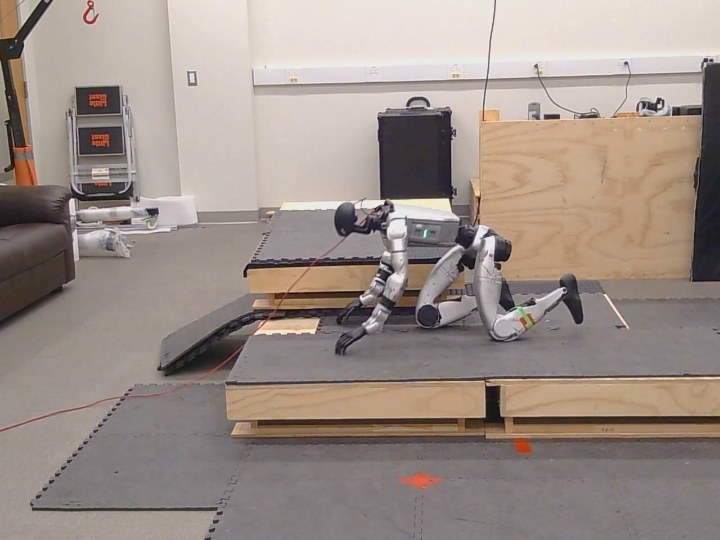}};
\node[anchor=north east,inner sep=0,outer sep=0,draw=white,line width=0.25pt] at (scene.north east) {\includegraphics[width=0.18623\linewidth,trim=465bp 341bp 0bp 0bp,clip]{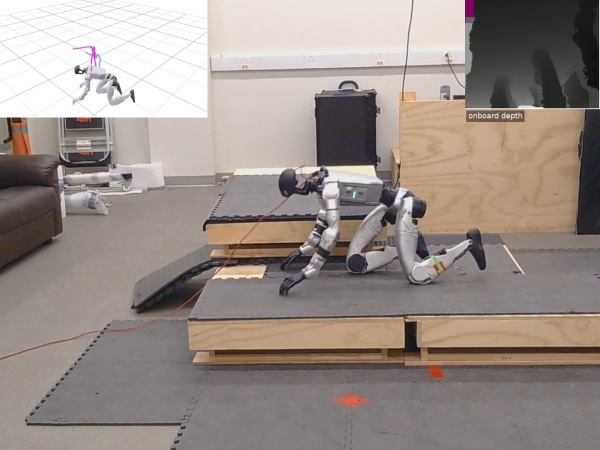}};
\end{tikzpicture}
\end{minipage}\hfill
\begin{minipage}[t]{0.247000\linewidth}
\centering
\begin{tikzpicture}
\node[inner sep=0,outer sep=0] (scene) {\includegraphics[width=\linewidth,trim=0bp 60bp 0bp 0bp,clip]{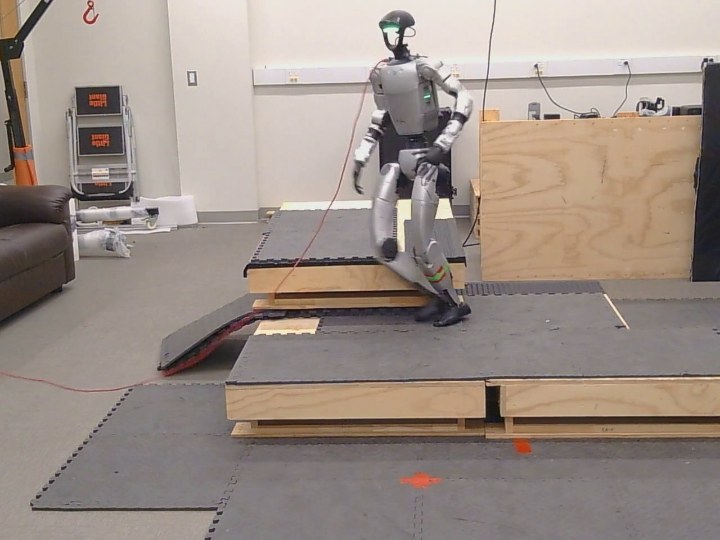}};
\node[anchor=north east,inner sep=0,outer sep=0,draw=white,line width=0.25pt] at (scene.north east) {\includegraphics[width=0.18623\linewidth,trim=465bp 341bp 0bp 0bp,clip]{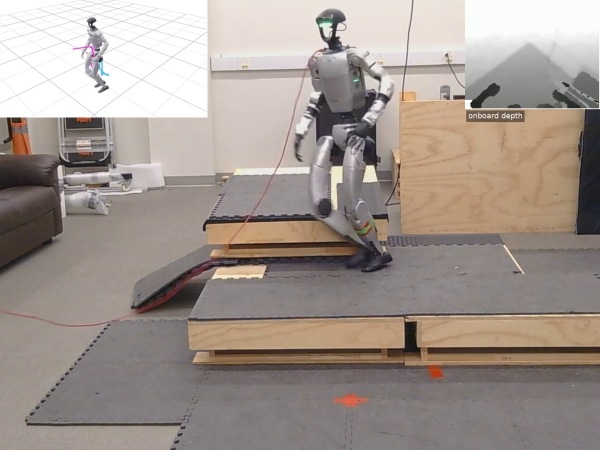}};
\end{tikzpicture}
\end{minipage}\par\nointerlineskip
\filmrail
\end{minipage}
\end{minipage}\par\vspace{3pt}
\begin{minipage}[t]{0.493\textwidth}
{\footnotesize\textbf{(B) Steered recovery on flat ground}}\par\vspace{1pt}\nointerlineskip
\begin{minipage}[c]{0.125\linewidth}
\centering
{\scriptsize\color{magenta!70!black} Guidance}\par
\includegraphics[width=\linewidth]{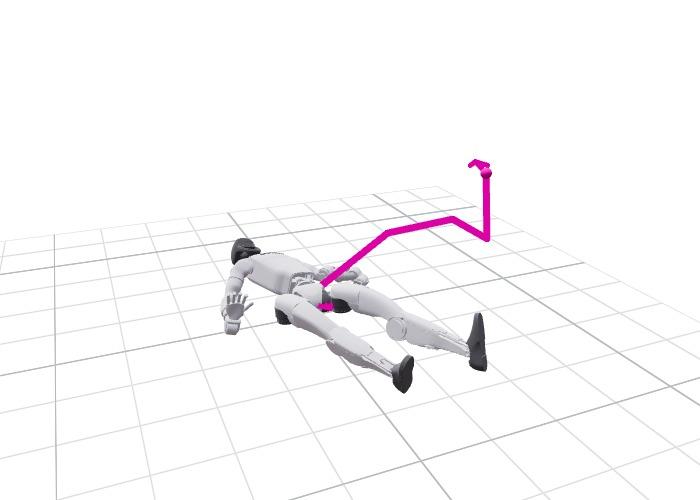}
\end{minipage}\hfill
\begin{minipage}[c]{0.865\linewidth}
\filmrail\par\nointerlineskip
\begin{minipage}[t]{0.247000\linewidth}
\centering
\begin{tikzpicture}
\node[inner sep=0,outer sep=0] (scene) {\includegraphics[width=\linewidth,trim=100bp 30bp 80bp 150bp,clip]{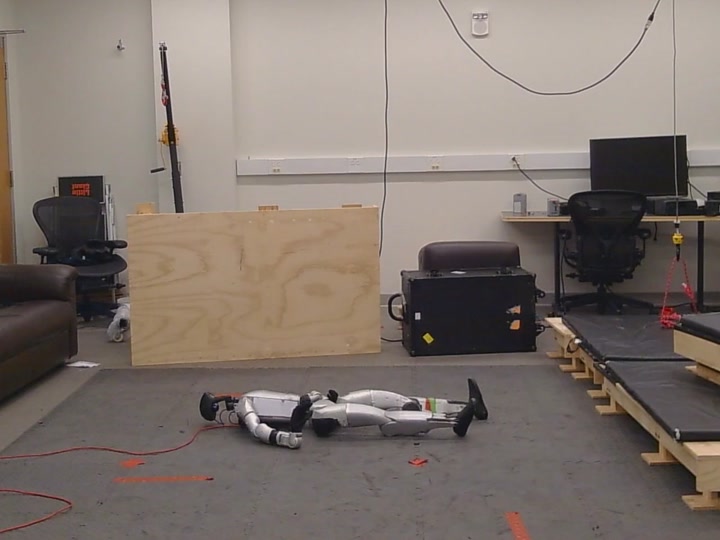}};
\node[anchor=north east,inner sep=0,outer sep=0,draw=white,line width=0.25pt] at (scene.north east) {\includegraphics[width=0.18623\linewidth]{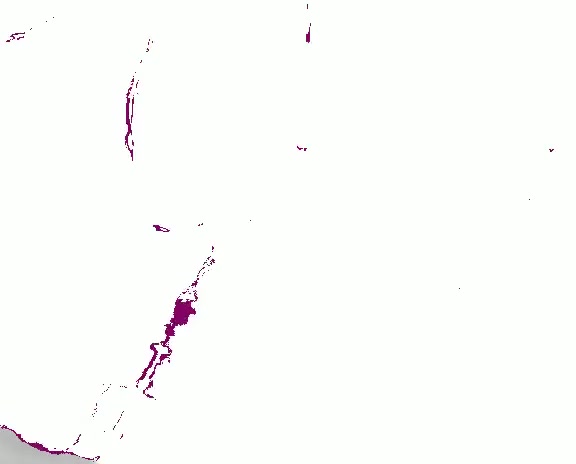}};
\end{tikzpicture}
\end{minipage}\hfill
\begin{minipage}[t]{0.247000\linewidth}
\centering
\begin{tikzpicture}
\node[inner sep=0,outer sep=0] (scene) {\includegraphics[width=\linewidth,trim=100bp 30bp 80bp 150bp,clip]{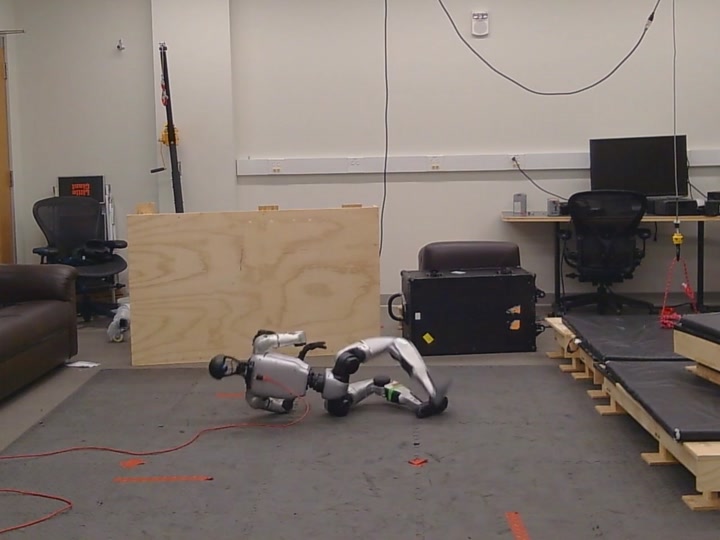}};
\node[anchor=north east,inner sep=0,outer sep=0,draw=white,line width=0.25pt] at (scene.north east) {\includegraphics[width=0.18623\linewidth]{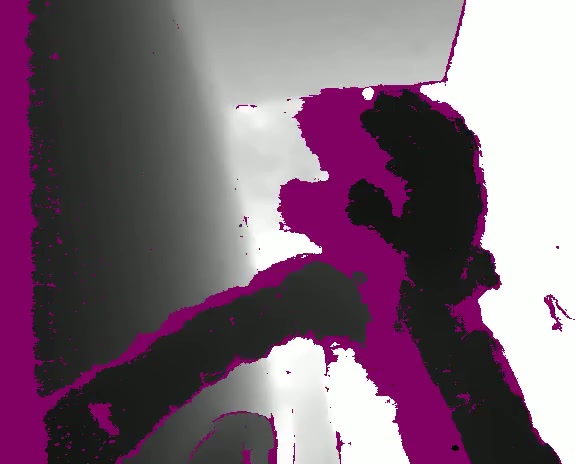}};
\end{tikzpicture}
\end{minipage}\hfill
\begin{minipage}[t]{0.247000\linewidth}
\centering
\begin{tikzpicture}
\node[inner sep=0,outer sep=0] (scene) {\includegraphics[width=\linewidth,trim=100bp 30bp 80bp 150bp,clip]{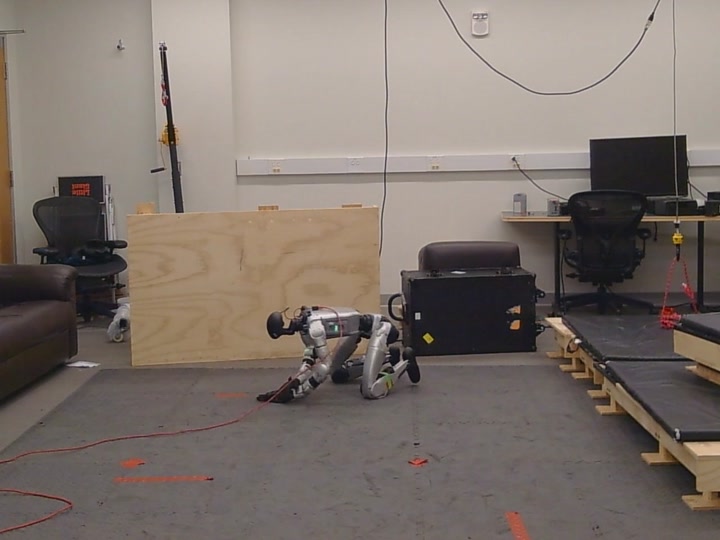}};
\node[anchor=north east,inner sep=0,outer sep=0,draw=white,line width=0.25pt] at (scene.north east) {\includegraphics[width=0.18623\linewidth]{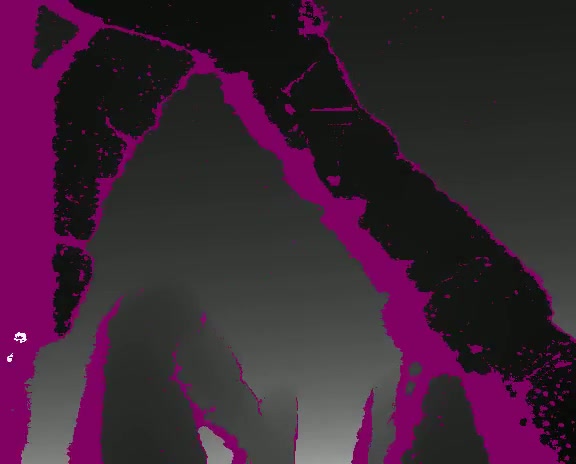}};
\end{tikzpicture}
\end{minipage}\hfill
\begin{minipage}[t]{0.247000\linewidth}
\centering
\begin{tikzpicture}
\node[inner sep=0,outer sep=0] (scene) {\includegraphics[width=\linewidth,trim=100bp 30bp 80bp 150bp,clip]{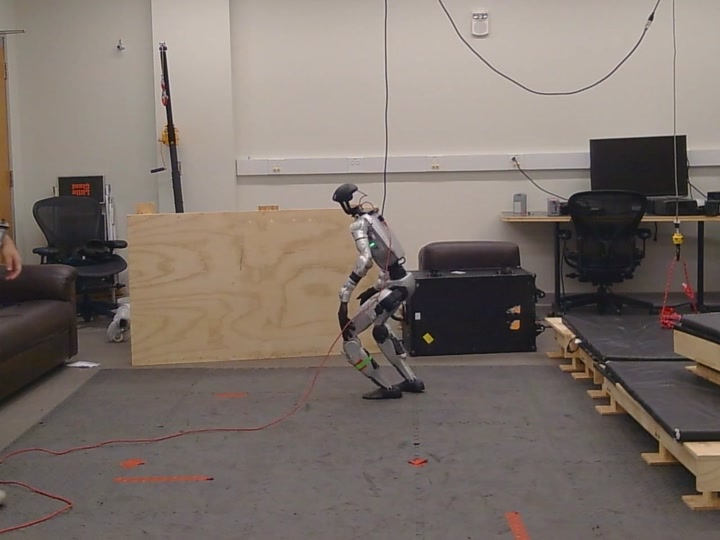}};
\node[anchor=north east,inner sep=0,outer sep=0,draw=white,line width=0.25pt] at (scene.north east) {\includegraphics[width=0.18623\linewidth]{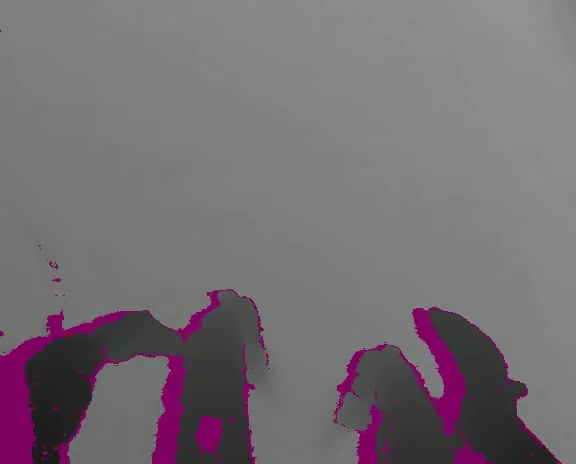}};
\end{tikzpicture}
\end{minipage}\par\nointerlineskip
\filmrail
\end{minipage}
\end{minipage}\hfill
\begin{minipage}[t]{0.493\textwidth}
{\footnotesize\textbf{(C) Steered recovery beside a wall}}\par\vspace{1pt}\nointerlineskip
\begin{minipage}[c]{0.125\linewidth}
\centering
{\scriptsize\color{magenta!70!black} Guidance}\par
\includegraphics[width=\linewidth]{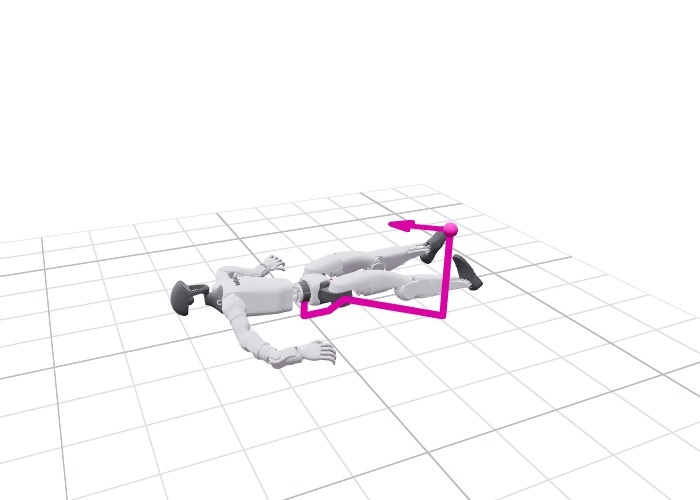}
\end{minipage}\hfill
\begin{minipage}[c]{0.865\linewidth}
\filmrail\par\nointerlineskip
\begin{minipage}[t]{0.247000\linewidth}
\centering
\begin{tikzpicture}
\node[inner sep=0,outer sep=0] (scene) {\includegraphics[width=\linewidth,trim=0bp 0bp 0bp 60bp,clip]{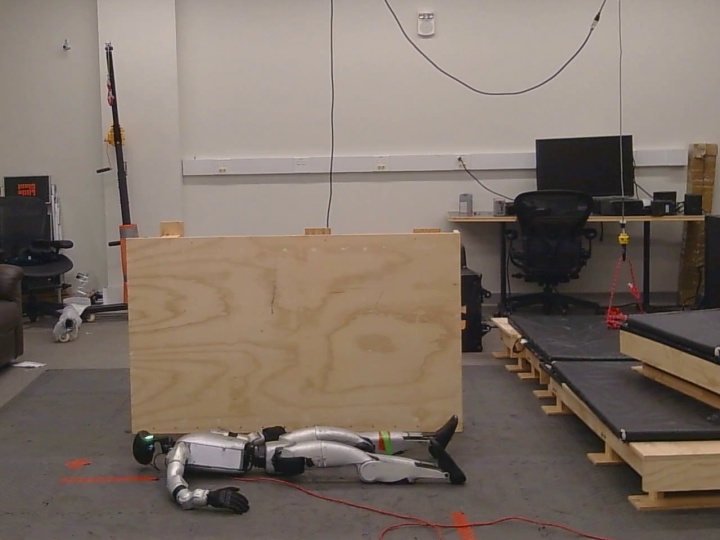}};
\node[anchor=north east,inner sep=0,outer sep=0,draw=white,line width=0.25pt] at (scene.north east) {\includegraphics[width=0.18623\linewidth]{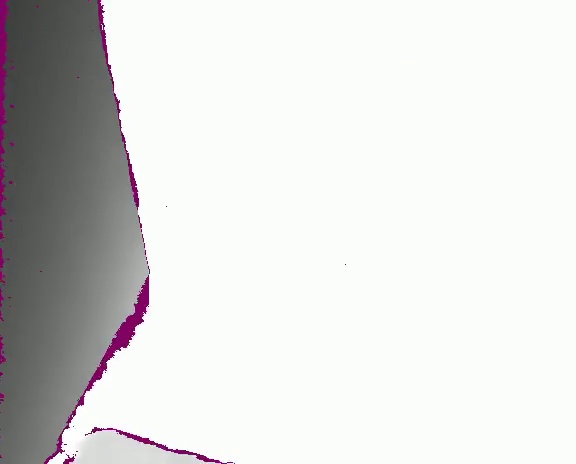}};
\end{tikzpicture}
\end{minipage}\hfill
\begin{minipage}[t]{0.247000\linewidth}
\centering
\begin{tikzpicture}
\node[inner sep=0,outer sep=0] (scene) {\includegraphics[width=\linewidth,trim=0bp 0bp 0bp 60bp,clip]{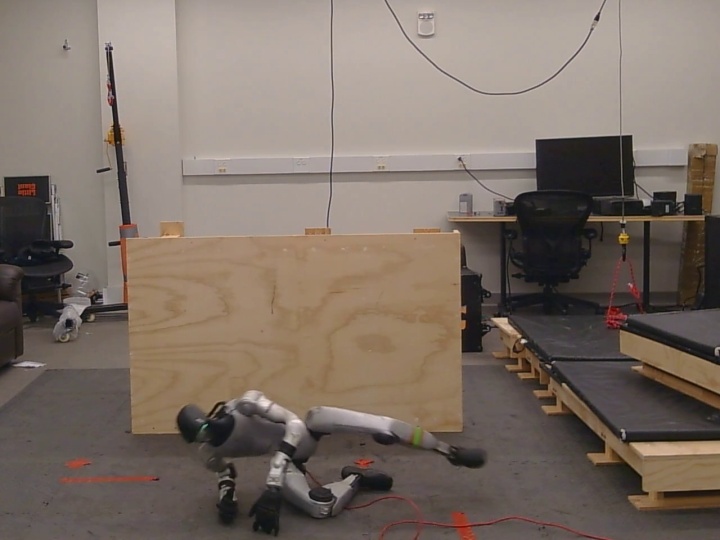}};
\node[anchor=north east,inner sep=0,outer sep=0,draw=white,line width=0.25pt] at (scene.north east) {\includegraphics[width=0.18623\linewidth]{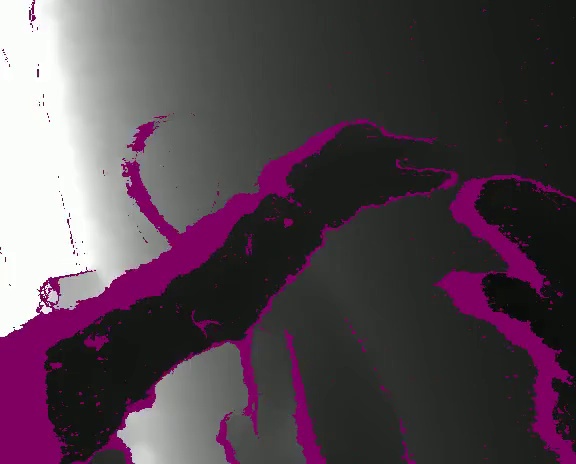}};
\end{tikzpicture}
\end{minipage}\hfill
\begin{minipage}[t]{0.247000\linewidth}
\centering
\begin{tikzpicture}
\node[inner sep=0,outer sep=0] (scene) {\includegraphics[width=\linewidth,trim=0bp 0bp 0bp 60bp,clip]{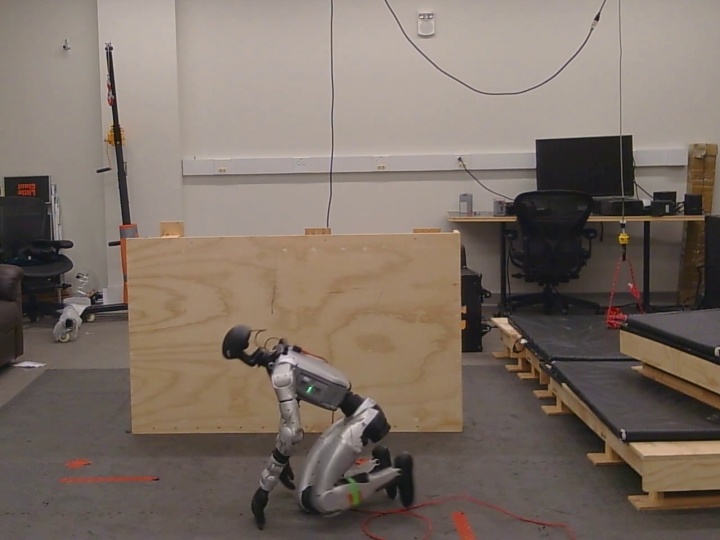}};
\node[anchor=north east,inner sep=0,outer sep=0,draw=white,line width=0.25pt] at (scene.north east) {\includegraphics[width=0.18623\linewidth]{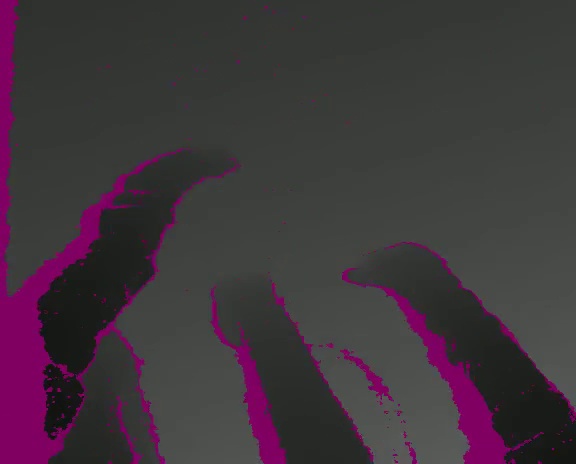}};
\end{tikzpicture}
\end{minipage}\hfill
\begin{minipage}[t]{0.247000\linewidth}
\centering
\begin{tikzpicture}
\node[inner sep=0,outer sep=0] (scene) {\includegraphics[width=\linewidth,trim=0bp 0bp 0bp 60bp,clip]{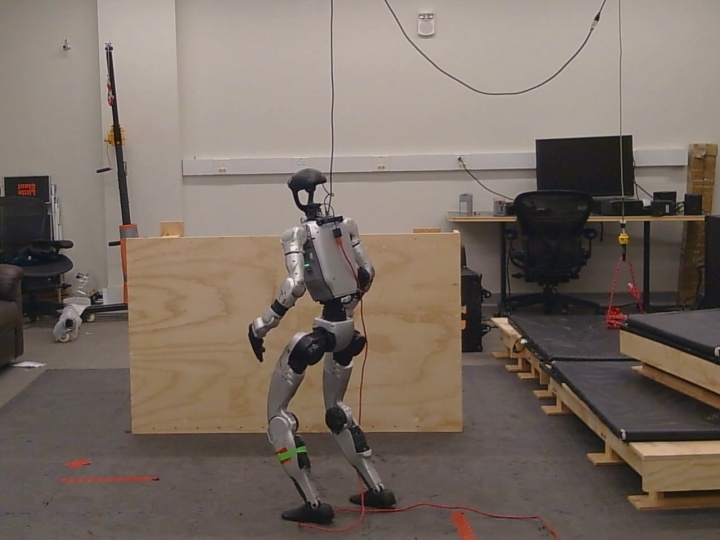}};
\node[anchor=north east,inner sep=0,outer sep=0,draw=white,line width=0.25pt] at (scene.north east) {\includegraphics[width=0.18623\linewidth]{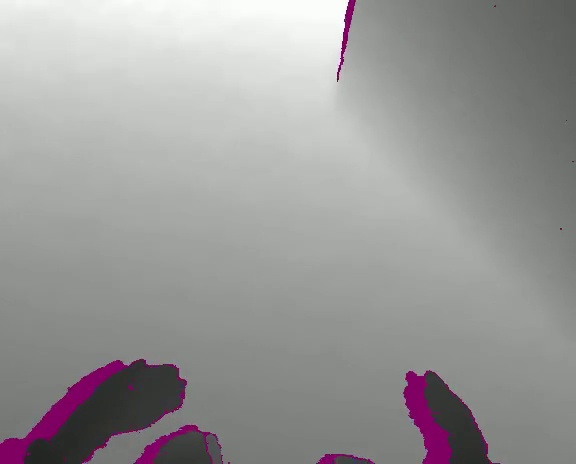}};
\end{tikzpicture}
\end{minipage}\par\nointerlineskip
\filmrail
\end{minipage}
\end{minipage}\par\vspace{3pt}
\begin{minipage}[t]{0.366\textwidth}
{\footnotesize\textbf{(D) Unguided (left)}}\par\vspace{1pt}\nointerlineskip
\filmrail\par\nointerlineskip
\begin{minipage}[t]{0.328000\linewidth}
\centering
\begin{tikzpicture}
\node[inner sep=0,outer sep=0] (scene) {\includegraphics[width=\linewidth,trim=25bp 45bp 215bp 135bp,clip]{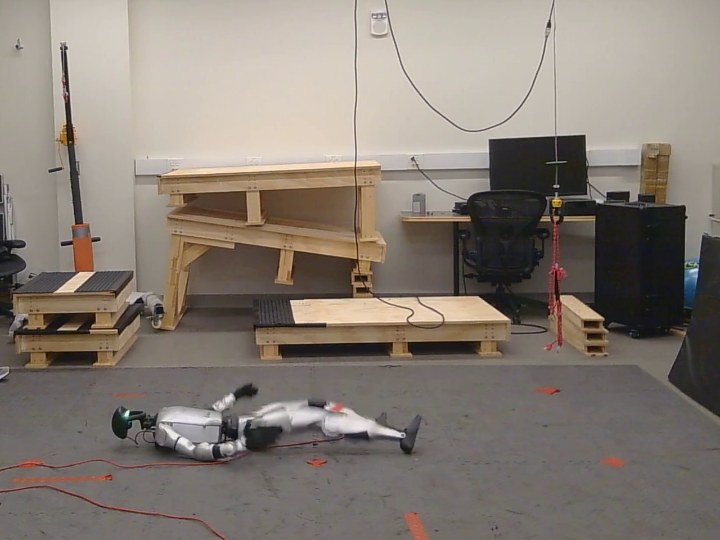}};
\node[anchor=north east,inner sep=0,outer sep=0,draw=white,line width=0.25pt] at (scene.north east) {\includegraphics[width=0.20243\linewidth,trim=0bp 341bp 465bp 0bp,clip]{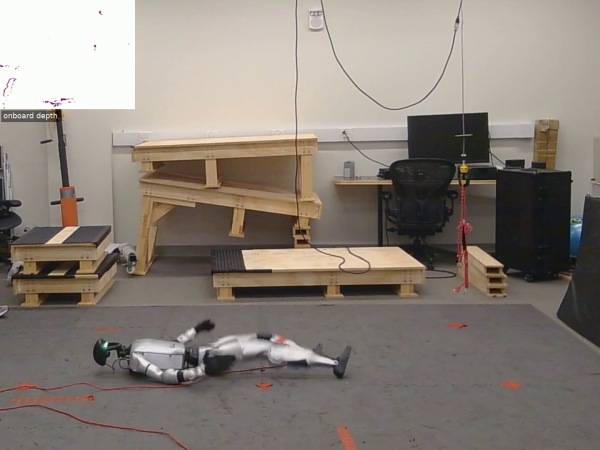}};
\end{tikzpicture}
\end{minipage}\hfill
\begin{minipage}[t]{0.328000\linewidth}
\centering
\begin{tikzpicture}
\node[inner sep=0,outer sep=0] (scene) {\includegraphics[width=\linewidth,trim=25bp 45bp 215bp 135bp,clip]{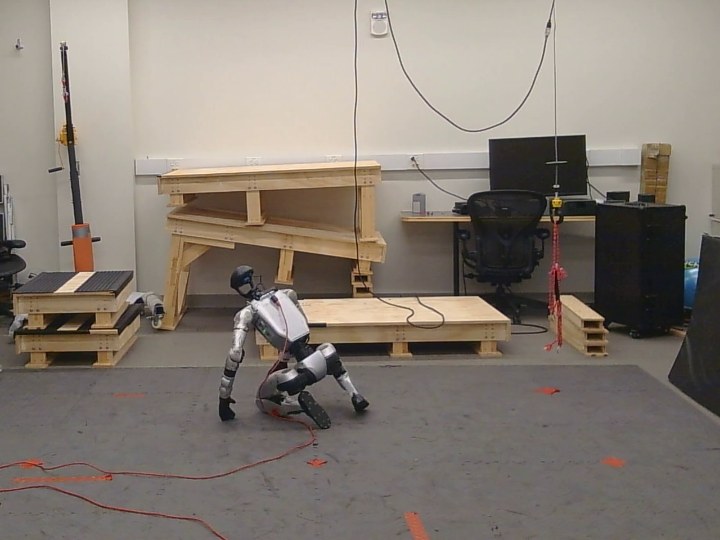}};
\node[anchor=north east,inner sep=0,outer sep=0,draw=white,line width=0.25pt] at (scene.north east) {\includegraphics[width=0.20243\linewidth,trim=0bp 341bp 465bp 0bp,clip]{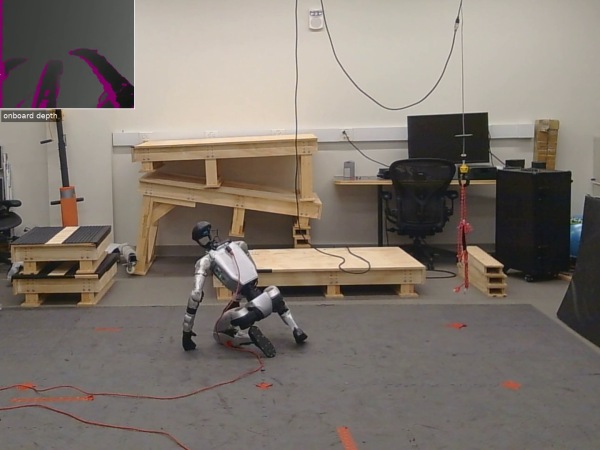}};
\end{tikzpicture}
\end{minipage}\hfill
\begin{minipage}[t]{0.328000\linewidth}
\centering
\begin{tikzpicture}
\node[inner sep=0,outer sep=0] (scene) {\includegraphics[width=\linewidth,trim=25bp 45bp 215bp 135bp,clip]{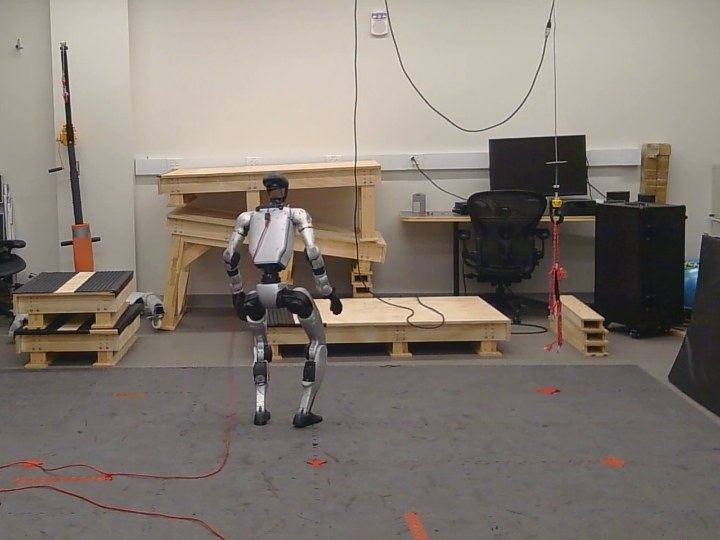}};
\node[anchor=north east,inner sep=0,outer sep=0,draw=white,line width=0.25pt] at (scene.north east) {\includegraphics[width=0.20243\linewidth,trim=0bp 341bp 465bp 0bp,clip]{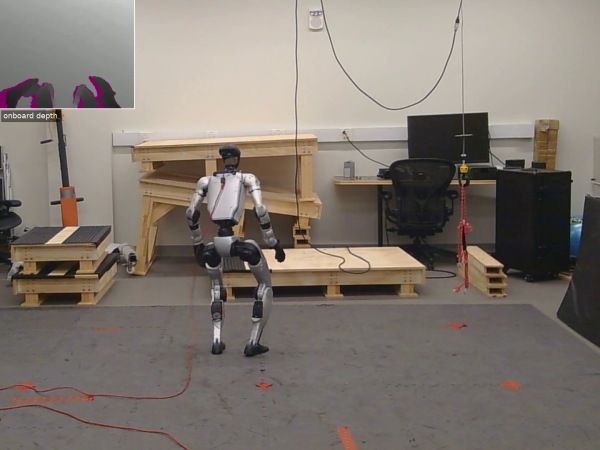}};
\end{tikzpicture}
\end{minipage}\par\nointerlineskip
\filmrail
\end{minipage}\hfill
\begin{minipage}[t]{0.242\textwidth}
{\footnotesize\textbf{(E) Unguided (in place)}}\par\vspace{1pt}\nointerlineskip
\filmrail\par\nointerlineskip
\begin{minipage}[t]{0.496\linewidth}
\centering
\begin{tikzpicture}
\node[inner sep=0,outer sep=0] (scene) {\includegraphics[width=\linewidth,trim=35bp 45bp 205bp 135bp,clip]{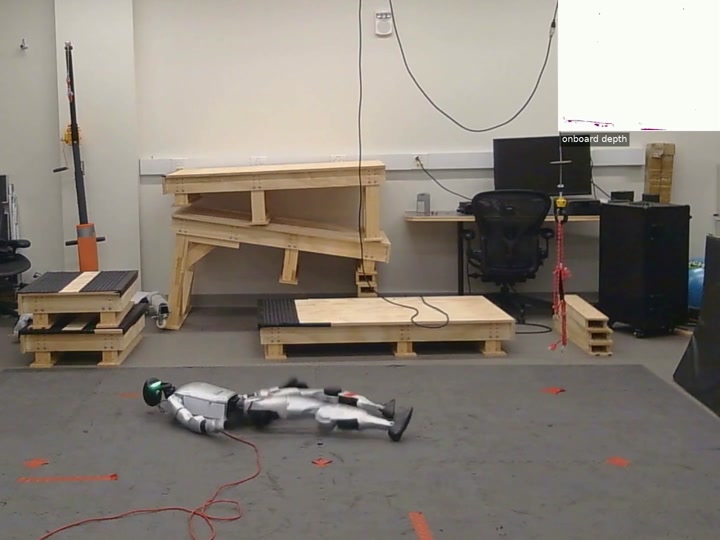}};
\node[anchor=center,inner sep=0,outer sep=0,opacity=0.72] at (scene.center) {\includegraphics[width=\linewidth,trim=35bp 45bp 205bp 135bp,clip]{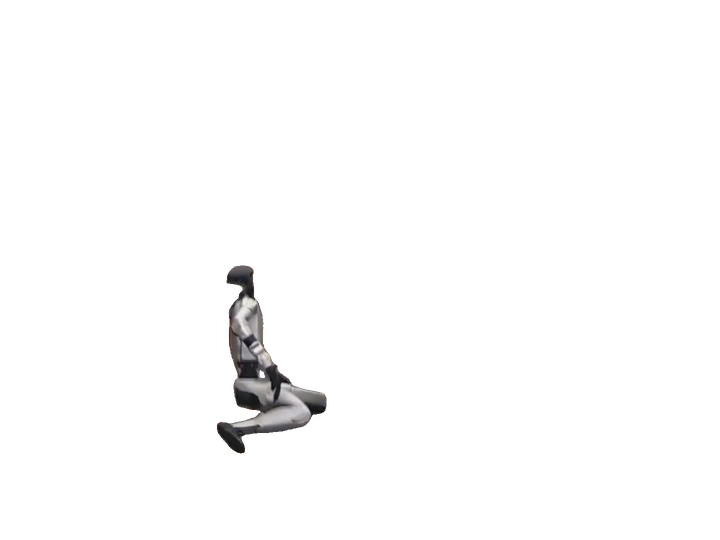}};
\node[anchor=north east,inner sep=0,outer sep=0,draw=white,line width=0.25pt] at (scene.north east) {\includegraphics[width=0.20243\linewidth]{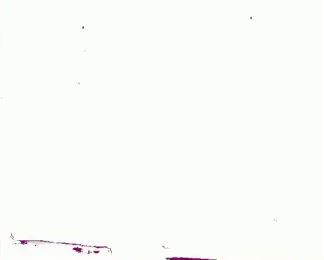}};
\end{tikzpicture}
\end{minipage}\hfill
\begin{minipage}[t]{0.496\linewidth}
\centering
\begin{tikzpicture}
\node[inner sep=0,outer sep=0] (scene) {\includegraphics[width=\linewidth,trim=35bp 45bp 205bp 135bp,clip]{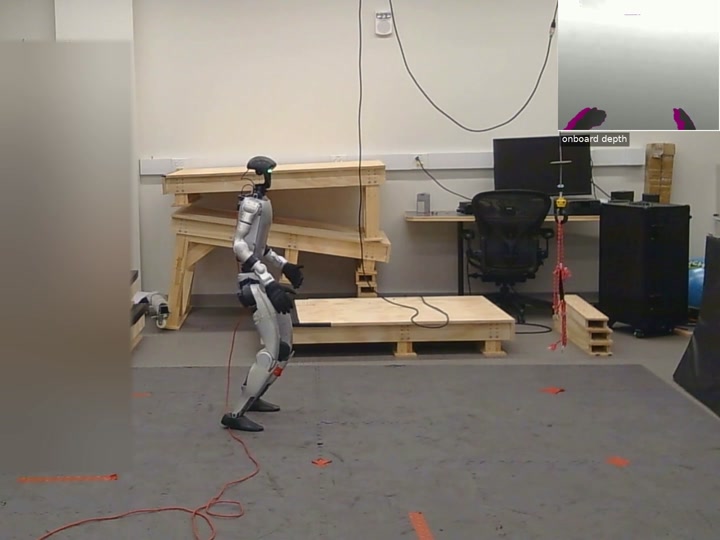}};
\node[anchor=north east,inner sep=0,outer sep=0,draw=white,line width=0.25pt] at (scene.north east) {\includegraphics[width=0.20243\linewidth]{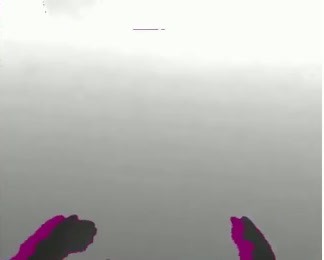}};
\end{tikzpicture}
\end{minipage}\par\nointerlineskip
\filmrail
\end{minipage}\hfill
\begin{minipage}[t]{0.366\textwidth}
{\footnotesize\textbf{(F) Unguided (right)}}\par\vspace{1pt}\nointerlineskip
\filmrail\par\nointerlineskip
\begin{minipage}[t]{0.328000\linewidth}
\centering
\begin{tikzpicture}
\node[inner sep=0,outer sep=0] (scene) {\includegraphics[width=\linewidth,trim=50bp 30bp 190bp 150bp,clip]{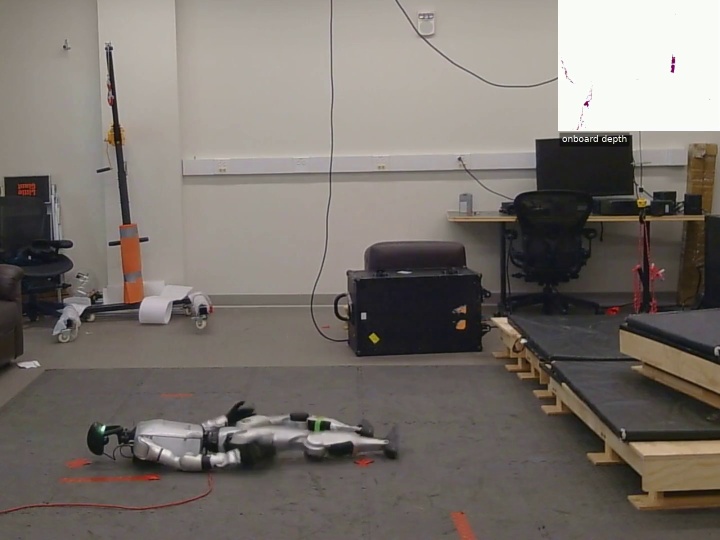}};
\node[anchor=north east,inner sep=0,outer sep=0,draw=white,line width=0.25pt] at (scene.north east) {\includegraphics[width=0.20243\linewidth,trim=465bp 341bp 0bp 0bp,clip]{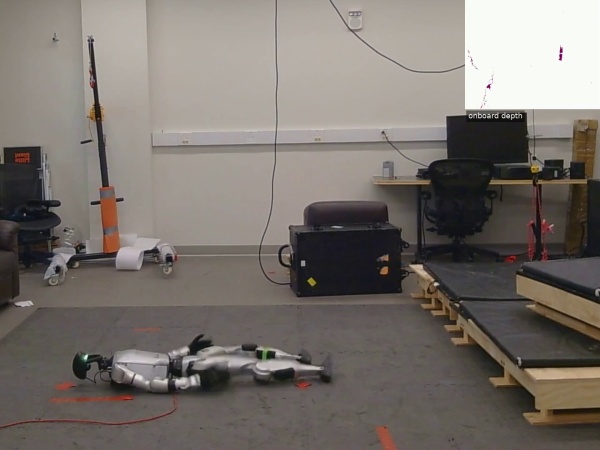}};
\end{tikzpicture}
\end{minipage}\hfill
\begin{minipage}[t]{0.328000\linewidth}
\centering
\begin{tikzpicture}
\node[inner sep=0,outer sep=0] (scene) {\includegraphics[width=\linewidth,trim=50bp 30bp 190bp 150bp,clip]{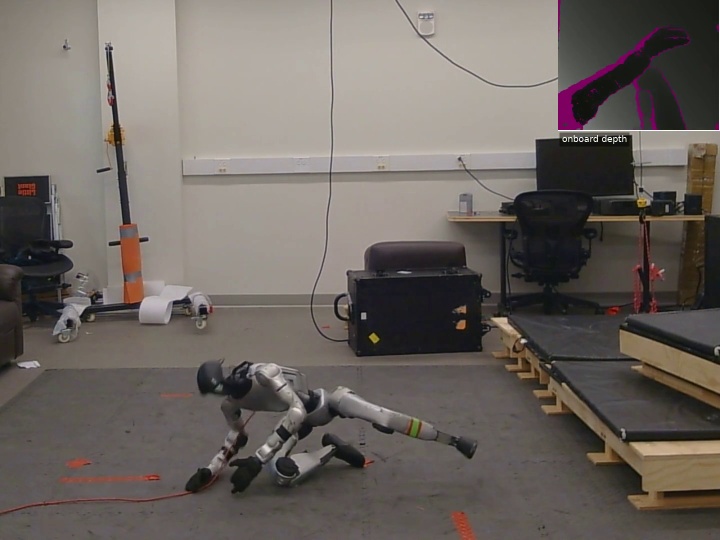}};
\node[anchor=north east,inner sep=0,outer sep=0,draw=white,line width=0.25pt] at (scene.north east) {\includegraphics[width=0.20243\linewidth,trim=465bp 341bp 0bp 0bp,clip]{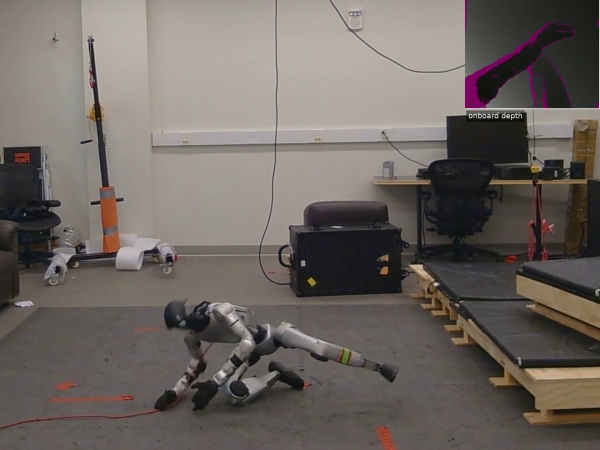}};
\end{tikzpicture}
\end{minipage}\hfill
\begin{minipage}[t]{0.328000\linewidth}
\centering
\begin{tikzpicture}
\node[inner sep=0,outer sep=0] (scene) {\includegraphics[width=\linewidth,trim=50bp 30bp 190bp 150bp,clip]{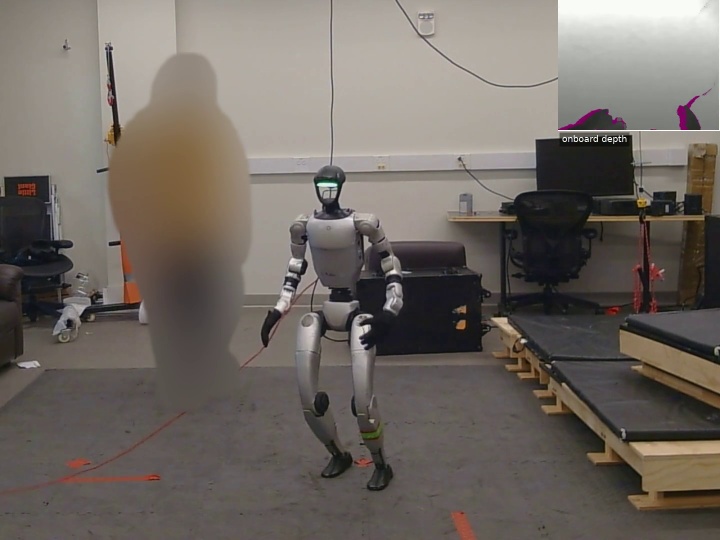}};
\node[anchor=north east,inner sep=0,outer sep=0,draw=white,line width=0.25pt] at (scene.north east) {\includegraphics[width=0.20243\linewidth,trim=465bp 341bp 0bp 0bp,clip]{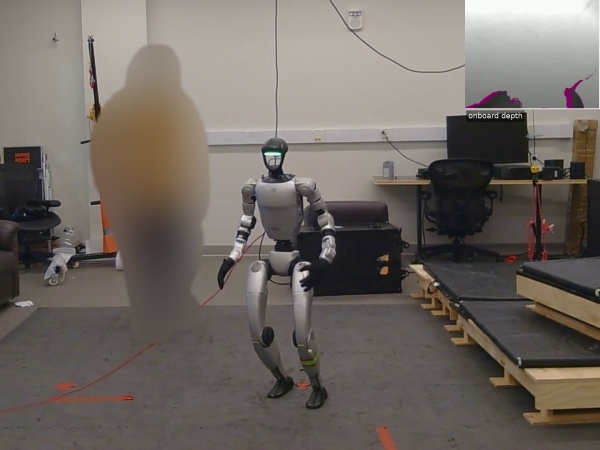}};
\end{tikzpicture}
\end{minipage}\par\nointerlineskip
\filmrail
\end{minipage}\par\vspace{3pt}
\begin{minipage}[t]{1\textwidth}
{\footnotesize\textbf{(G) Recovery on uneven supports}}\par\vspace{1pt}\nointerlineskip
\filmrail\par\nointerlineskip
\begin{minipage}[t]{0.247000\linewidth}
\centering
\begin{tikzpicture}
\node[inner sep=0,outer sep=0] (scene) {\includegraphics[width=\linewidth,trim=0bp 60bp 0bp 100bp,clip]{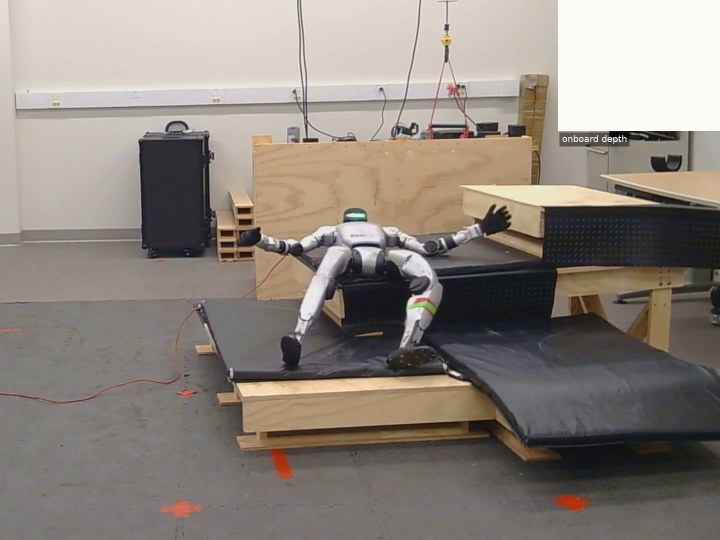}};
\node[anchor=north east,inner sep=0,outer sep=0,draw=white,line width=0.25pt] at (scene.north east) {\includegraphics[width=0.225\linewidth,trim=558bp 390bp 0bp 0bp,clip]{figures/hardware_gallery/terrain_b/scene_01.jpg}};
\end{tikzpicture}
\end{minipage}\hfill
\begin{minipage}[t]{0.247000\linewidth}
\centering
\begin{tikzpicture}
\node[inner sep=0,outer sep=0] (scene) {\includegraphics[width=\linewidth,trim=0bp 60bp 0bp 100bp,clip]{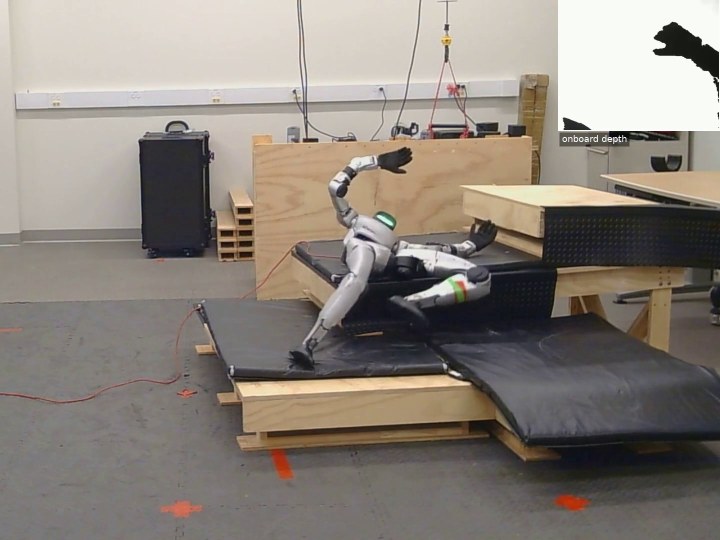}};
\node[anchor=north east,inner sep=0,outer sep=0,draw=white,line width=0.25pt] at (scene.north east) {\includegraphics[width=0.225\linewidth,trim=558bp 390bp 0bp 0bp,clip]{figures/hardware_gallery/terrain_b/scene_02.jpg}};
\end{tikzpicture}
\end{minipage}\hfill
\begin{minipage}[t]{0.247000\linewidth}
\centering
\begin{tikzpicture}
\node[inner sep=0,outer sep=0] (scene) {\includegraphics[width=\linewidth,trim=0bp 60bp 0bp 100bp,clip]{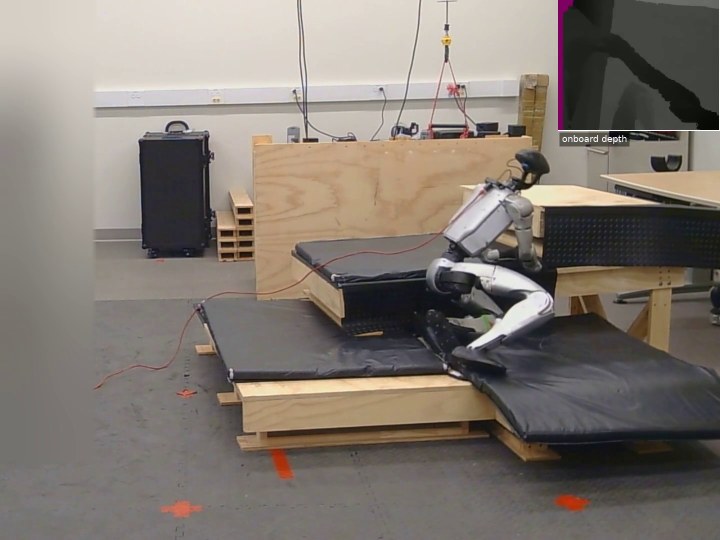}};
\node[anchor=north east,inner sep=0,outer sep=0,draw=white,line width=0.25pt] at (scene.north east) {\includegraphics[width=0.225\linewidth,trim=558bp 390bp 0bp 0bp,clip]{figures/hardware_gallery/terrain_b/scene_03.jpg}};
\end{tikzpicture}
\end{minipage}\hfill
\begin{minipage}[t]{0.247000\linewidth}
\centering
\begin{tikzpicture}
\node[inner sep=0,outer sep=0] (scene) {\includegraphics[width=\linewidth,trim=0bp 150bp 0bp 10bp,clip]{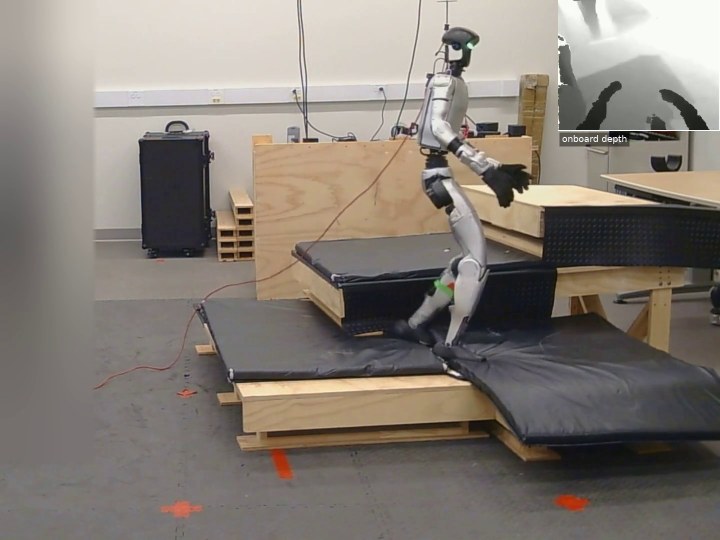}};
\node[anchor=north east,inner sep=0,outer sep=0,draw=white,line width=0.25pt] at (scene.north east) {\includegraphics[width=0.225\linewidth,trim=558bp 390bp 0bp 0bp,clip]{figures/hardware_gallery/terrain_b/scene_04.jpg}};
\end{tikzpicture}
\end{minipage}\par\nointerlineskip
\filmrail
\end{minipage}\par\nointerlineskip
\caption{\textbf{Real-world recovery.} Steered recovery (A--C), unguided
supine recoveries (D--F), and recovery on uneven supports (G).
Magenta paths show guidance; insets show onboard depth. Frames advance
in time from left to right within each panel.}
\label{fig:real-hardware-gallery}
\label{fig:real-steering}
\label{fig:real-stairs}
\par\vspace{3pt}\nointerlineskip
\begin{minipage}[t]{\columnwidth}
\vspace{0pt}
\centering
\includegraphics[width=\columnwidth]{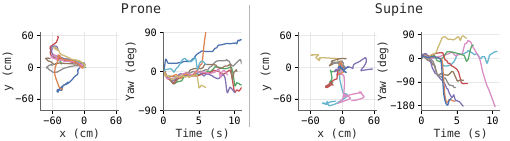}
\par\nointerlineskip
\caption{\textbf{Unguided flat-ground recoveries.} Colors identify trials;
motion is relative to the initial pose ($x$ forward, $y$ left).}
\label{fig:real-pelvis-motion}
\end{minipage}\hfill
\begin{minipage}[t]{\columnwidth}
\vspace{0pt}
\centering
\includegraphics[width=\columnwidth]{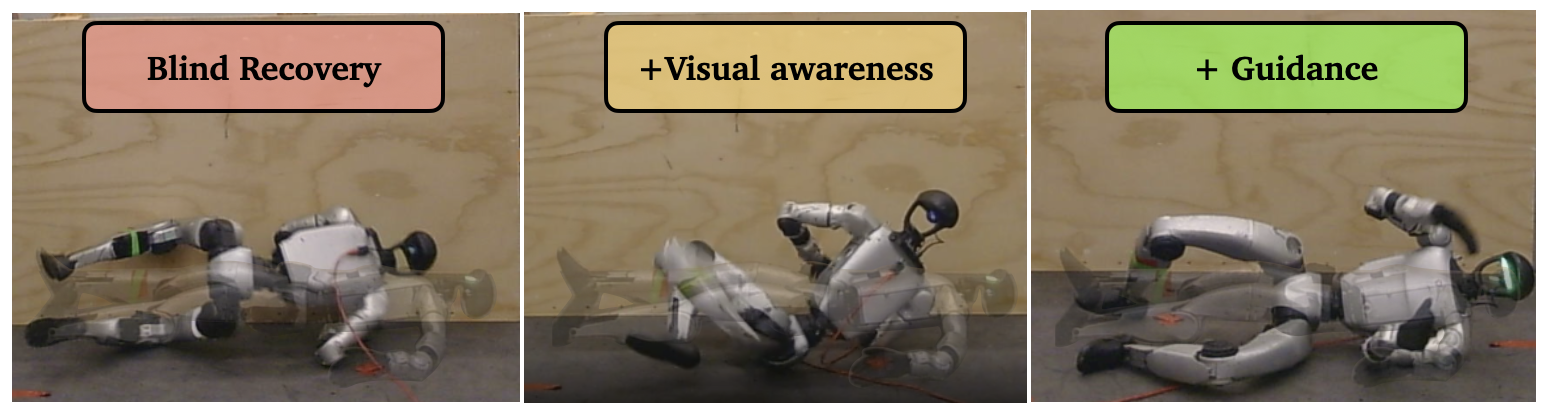}
\par\nointerlineskip
\caption{\textbf{Recovery beside a wall.} Blind recovery collides with the wall.
Vision changes the route, and guidance can favor the open side.}
\label{fig:real-blind-visual-guidance}

\end{minipage}
\end{figure*}

\textbf{Task-directed recovery.}
Table~\ref{tab:eval-flat-steering} shows that task guidance shapes recovery
while maintaining high success across the tested terrains. From the same
initial configuration, the robot follows different headings, rise speeds, and paths
(Figs.~\ref{fig:sim-steering-examples} and~\ref{fig:eval-steering}).
Slower targets (later rise times) have lower errors and higher success.
Faster rises may be harder to control as momentum and support contacts
change rapidly.

\textbf{Steering on uneven terrain.}
A requested motion must also fit the support available on the terrain.
Changing the path or heading can change the hand, foot, and body contacts
used to rise, making these choices harder to realize on uneven ground.
In Table~\ref{tab:eval-flat-steering}, heading errors increase on stairs and
the slope, and rise-time errors are largest on the slope. The clearest
limitation appears in path steering: tracking errors remain close to
flat-ground values, but recovery succeeds less often. Accurate path tracking alone
does not ensure a stable recovery on uneven terrain.

\subsection{Role of Visual Feedback}
\label{sec:eval-depth}

Visual feedback can inform motion selection and execution.
We compare variants trained without depth in the planner alone
(\emph{blind planner}) or in both the planner and decoder
(\emph{both blind}). We test each variant on uneven terrain using 120
unguided trials from matched initial configurations.

In Table~\ref{tab:depth-perception}, removing planner depth reduces recovery
success even when the decoder still observes the terrain. This suggests
visual feedback during execution cannot fully compensate for planning
without scene information. Removing decoder depth further reduces
success. The two blind variants show greater displacement, suggesting
more drift and difficulty settling, as well as higher peak torque and
joint-limit excess. These results support visual feedback in both planning
and execution.

\subsection{State-First Generation and Replanning}
\label{sec:eval-joint}

We examine whether conditioning motion latents on steered states helps the
robot track task objectives while completing recovery. We compare state-first
generation with joint diffusion using the same decoder, training budget,
initial configurations, targets, and guidance strengths, with
120 trials per objective, model, and replanning interval. We test heading,
speed, and path objectives at replanning intervals of 0.2, 0.5, and 1.0\,s.
Each objective uses three initial configurations selected from flat ground,
stepped terrain, stairs, and rubble.
We report MAE for heading and rise time, and RMSE for height, position,
and path. Speed steering uses rise time in
Table~\ref{tab:eval-flat-steering} and height profiles in
Table~\ref{tab:steering-replanning}.

\textbf{State-first and joint generation.}
In Table~\ref{tab:steering-replanning}, state-first generation achieves lower
tracking errors for all three objectives and higher recovery success at the
default replanning interval. Its benefit extends beyond tracking:
at the longest interval, heading errors are similar, yet state-first
completes recovery more often. These results support steering predicted
states before generating motion latents for execution.

\textbf{Replanning interval.}
Longer intervals increase tracking errors for both models, while
state-first heading success increases.
Frequent replanning improves tracking without consistently
improving recovery success.

\subsection[Real-World Experiments]{Real-World Experiments}
\label{sec:realworld}

\textbf{Experimental setup.}
We examine recovery diversity,
guidance, and terrain awareness on a 29-DoF Unitree G1 equipped with
proprioception and an egocentric depth camera. Inference runs offboard on an RTX~4090,
with asynchronous planning and a 50\,Hz decoder.
We use DLIO~\cite{chen2023dlio} to estimate pelvis motion from LiDAR and IMU measurements.

\begin{table}[t]
\centering\footnotesize\setlength{\tabcolsep}{4pt}
\renewcommand{\arraystretch}{1.0}
\caption{Real-world recovery outcomes}
\label{tab:real-recovery}
\begin{tabular*}{\columnwidth}{@{\extracolsep{\fill}}lclc@{}}
\toprule
\multicolumn{2}{c}{\textit{Unguided recovery}} &
\multicolumn{2}{c}{\textit{Steered recovery}} \\
\cmidrule(r){1-2}\cmidrule(l){3-4}
\textbf{Setup} & \textbf{Success} & \textbf{Setup} & \textbf{Success} \\
\midrule
Flat (prone) & 10/10 & Flat (prone) & 4/5 \\
Flat (supine) & 10/10 & Flat (supine) & 4/5 \\
Staggered steps & 9/10 & Near wall & 5/5 \\
\bottomrule
\end{tabular*}
\end{table}

\textbf{Recovery diversity.} From the same supine posture, the robot can turn
left, rise in place, or turn right as it returns to standing
(Fig.~\ref{fig:real-hardware-gallery}D--F). We obtain these alternatives by
resetting to the same commanded initial pose and resampling diffusion noise.
Repeated trials from prone and supine starts follow different paths and reach
different final headings (Fig.~\ref{fig:real-pelvis-motion}), with all unguided
flat-ground trials recovering successfully (Table~\ref{tab:real-recovery}).
These results show that the learned prior supports multiple successful recovery
motions from a shared starting posture.\looseness=-1\relax

\begingroup\clubpenalty=10000
\textbf{Steered recovery.} Guidance allows a requested direction or route to
shape how the robot returns to standing. We demonstrate directional steering
from prone and supine starts on flat ground and route steering beside a wall
and across raised platforms (Fig.~\ref{fig:real-hardware-gallery}A--C).
Most flat-ground steering trials and all near-wall trials recover successfully
(Table~\ref{tab:real-recovery}). These examples illustrate how the same recovery
prior can support different task preferences through guidance at inference time.
\par\endgroup

The wall example illustrates the complementary roles of perception and
guidance (Fig.~\ref{fig:real-blind-visual-guidance}).
From the same nominal pose, blind recovery
rotates into the wall and collides with it. With depth observations, the
robot initially turns toward the wall before changing direction toward the
unobstructed side. With guidance, it moves toward that side from the start.
Perception supports adaptation to the nearby obstacle, while guidance expresses
the preferred recovery direction.

\textbf{Terrain awareness.} On staggered steps, the robot adjusts hand and foot support
as it rises between levels (Fig.~\ref{fig:real-hardware-gallery}G), illustrating
how recovery accommodates uneven contacts. Unguided recovery succeeds in most
trials across the tested step arrangements and initial configurations
(Table~\ref{tab:real-recovery}). Steering introduces an additional challenge:
a requested direction or path may compete with the contacts needed to rise.
The feedback decoder must then reconcile the steered motion intent with
available support, highlighting the role of physical realization in recovery.

\FloatBarrier
\section{Conclusion and Future Work}
\label{sec:conclusion}
We present \sname, an approach to perceptive humanoid fall recovery that combines
a multimodal recovery repertoire with test-time steering. By steering predicted
motion intent before generating motion latents for feedback execution, our
visual planner lets task objectives shape recovery without retraining.
Simulation experiments demonstrate heading, speed, and path steering,
while real-world demonstrations illustrate perceptive and steerable recovery.
Comparisons in simulation support visual feedback in both
planning and execution, and show that state-first generation tracks objectives
more accurately. On uneven terrain, however, accurate path tracking does not always
lead to successful recovery. Future work will address this gap between
following the requested motion and maintaining the support needed to rise,
and extend planning to moving objects and external disturbances.

\FloatBarrier
\bibliographystyle{ieeetr}
\bibliography{arxiv_version/references}

\end{document}